\documentclass[
  nojss]{jss}

\usepackage[utf8]{inputenc}

\author{
Gilles Kratzer\\Zurich University \And Fraser Iain Lewis\\Danone Nutricia Research \AND Arianna Comin\\National Veterinary Institute \And Marta Pittavino\\Geneva University \And Reinhard Furrer\\Zurich University
}
\title{Additive Bayesian Network Modelling with the \proglang{R} Package
\pkg{abn}}

\Plainauthor{Gilles Kratzer, Fraser Iain Lewis, Arianna Comin, Marta Pittavino, Reinhard Furrer}
\Plaintitle{Additive Bayesian Network Modelling with the R Package abn}
\Shorttitle{Additive Bayesian Network Modelling}

\Abstract{
The \proglang{R} package \pkg{abn} is designed to fit additive Bayesian
models to observational datasets. It contains routines to score Bayesian
networks based on Bayesian or information theoretic formulations of
generalized linear models. It is equipped with exact search and greedy
search algorithms to select the best network. It supports a possible
blend of continuous, discrete and count data and input of prior
knowledge at a structural level. The Bayesian implementation supports
random effects to control for one-layer clustering. In this paper, we
give an overview of the methodology and illustrate the package's
functionalities using a veterinary dataset about respiratory diseases in
commercial swine production.
}

\Keywords{structure learning, graphical models, greedy search, exact search, scoring algorithm, GLM, graph theory, \proglang{R}}
\Plainkeywords{structure learning, graphical models, greedy and exact search, scoring algorithm, GLM, graph theory, R}

\Address{
    Gilles Kratzer\\
  Zurich University\\
  Department of Mathematics, Winterthurerstrasse 190, CH-8057 Zürich,
  Switzerland\\
  E-mail: \email{gilles.kratzer@math.uzh.ch}\\
  URL: \url{http://r-bayesian-networks.org/}\\~\\
          }

\usepackage{thumbpdf,lmodern} \usepackage{amsmath} \usepackage{amssymb} \usepackage{booktabs} \usepackage{bm} \usepackage{subfig} \usepackage{algorithm2e} \usepackage{float}              

\begin{document}

\hypertarget{sec:intro}{%
\section{Introduction}\label{sec:intro}}

Bayesian network (BN) modelling has an impressive track record in
analysing systems epidemiology datasets
\citep{cornet2016bayesian, mccormick2014frequent, hartnack2019additive},
especially in veterinary epidemiology
\citep{mccormick2013using, ludwig2013identifying, firestone2014applying, cornet2016bayesian, hartnack2016attitudes, pittavino2017comparison, RUCHTI2018, ruchti2019progression, comin}.
It is a particularly well-suited approach to better understand the
underlying structure of data when scientific understanding of the data
is at an early stage. BN modelling is designed to sort out directly from
indirectly related variables and offers a far richer modelling framework
than classical approaches in epidemiology like, e.g., regression
techniques or extensions thereof. In contrast to structural equation
modelling \citep{hair1998multivariate}, which requires expert knowledge
to design the model, the Additive Bayesian Network (ABN) method is a
data-driven approach \citep{lewis2013improving, kratzerabn}. It does not
rely on expert knowledge, but it can possibly incorporate it. Contrary
to other scoring approaches, it can handle a blend of variables
distribution. Thanks to its formulation, the ABN method supports
variables adjustment and can control for clustering. Within the same
framework, the network can be scored and the effect size estimated. This
paper aimes at presenting the actual \proglang{R} implementation of the
ABN methodology using a simple case study. An older package vignette
exists that targets presenting more detailed code.

\proglang{R} as an open-source, reliable and easy-to-use environment for
statistical computing. It is very popular in the epidemiological
community \citep{R-Core-Team:2017aa}. In this paper, ABN refers to the
methodology and \pkg{abn} refers to the \proglang{R} package.

The aim of the \proglang{R} package \pkg{abn} is to provide researchers
free implementation of a set of functions to score, select, analyse and
report an ABN modelling. The main functionalities of the package are
Bayesian, information theoretic scoring, exact and greey search
algorithms. The \proglang{R} package \pkg{abn} is also equipped with
ancillary functions to simulate and manipulate ABN models. The package
is available through the Comprehensive \proglang{R} Archive Network
(CRAN) at \url{https://CRAN.R-project.org/package=abn}. The unique
feature of the \proglang{R} package \pkg{abn} compared to other network
modelling \proglang{R} packages is the Bayesian-based scoring system,
which makes ABN methodology theoretically sound and computationally
competitive thanks to the internal use of iteratively nested Laplace
approximations (\proglang{INLA}) (\citealp{rue2009approximate},
available at \url{www.r-inla.org/download}). From an applied
perspective, a regression framework is suitable for analyses that target
data modelisation and seek to report insight of observational data. A
byproduct of the Bayesian regression framework used by the \proglang{R}
package \pkg{abn} is the possibility of adjusting an analysis for
clustering, which is a common concern in systems epidemiology and not
possible in other network modelling \proglang{R} packages. Beyond its
nice theoretical framework, the \proglang{R} package \pkg{abn} is
equipped with the only \proglang{R} implementation of an exact search
based on dynamical programming that targets BN modelling
\citep{koivisto2004exact}.

The structure of this paper is as follows: First, we finish this section
with a short motivating example and alternative approaches for modelling
BNs. Then, Section \ref{sec:method} describes the comprehensive ABN
methodology. Section \ref{sec:pkg} lists and describes the
functionalities of the \proglang{R} package. It includes simulation
studies for comparing the efficiency of the implemented scores. Section
\ref{sec:cs} presents a case study of an ABN modelling using data from
the field of veterinary epidemiology. We conclude and summarize the
article in Section \ref{sec:summary}. The appendix provides more
technical details.

\hypertarget{sec:motivatingexample}{%
\subsection{Motivating example}\label{sec:motivatingexample}}

We start illustrating the main functionality of the \proglang{R} package
\pkg{abn} by analyzing the so-called \texttt{asia} dataset provided by
the \proglang{R} package \pkg{bnlearn} \citep{Scutari:2010aa}. It is a
small synthetic dataset from \cite{laurits:88} about lung diseases
(tuberculosis, lung cancer and bronchitis) and visits to Asia. In total,
the dataset consists of eight binary variables. As this dataset has been
used in various illustrations, we do not present it here in detail, nor
do we perform an initial exploratory data analysis. We assume all
necessary packages are installed and we start with loading the data. The
advantage of using the \proglang{R} package \pkg{abn} over a classical
approach is that we take all data into account without making an a
priori choice of what are the independent and what is the dependent
variables.

\begin{CodeChunk}

\begin{CodeInput}
R> library("abn")
R> data("asia", package = "bnlearn")
R> colnames(asia) <- c("Asia", "Smoking", "Tuberculosis", 
+      "LungCancer", "Bronchitis", "Either", "XRay", "Dyspnea")
\end{CodeInput}
\end{CodeChunk}

We now determine the relationship between the data. A more theoretical
and more detailed explanation is given in Sections \ref{sec:method} and
\ref{sec:pkg}.

\begin{CodeChunk}

\begin{CodeInput}
R> distrib <- as.list(rep("binomial", 8))
R> names(distrib) <- names(asia)
R> 
R> mycache <- buildscorecache(data.df = asia, data.dist = distrib, 
+      max.parents = 4)
R> mp.dag <- mostprobable(score.cache = mycache)
\end{CodeInput}
\end{CodeChunk}

At this time, we have constructed the underlying relationship of the
variables. In our context, we also speak of we have learned about the
structure. In a next step, we estimate the parameters describing the
relationship between the variables, estimating the odds ratios of the
conditional effects.

\begin{CodeChunk}

\begin{CodeInput}
R> fabn <- fitabn(object = mp.dag, create.graph = TRUE)
R> require(Rgraphviz)
R> plot(fabn$abnDag)
\end{CodeInput}
\end{CodeChunk}

\begin{figure}[!ht] 
  \centering
\includegraphics[width=0.8\textwidth]{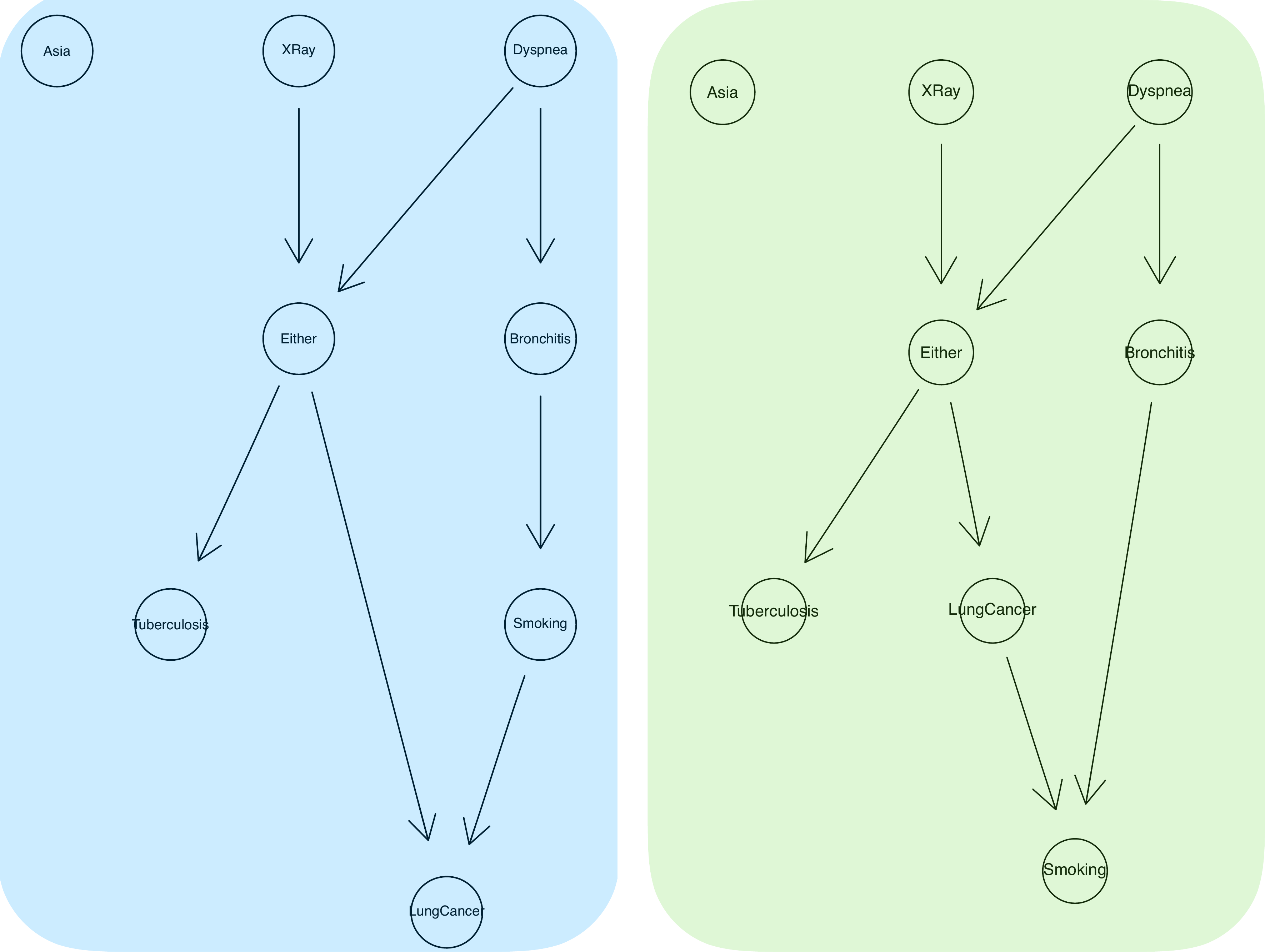}
  \caption{Network for \code{asia} dataset. Left: {\color{blue} unconstrained}, right: {\color{green} constrained}. The constrain is on the direction of the link between LungCancer and smoking nodes.} 
  \label{fig:asia}
\end{figure}

This example highlights that the \proglang{R} package \pkg{abn} does not
model causality. However, it is possible to structurally constrain the
modelling procedure in retaining or banning structures between the
variables.

\begin{CodeChunk}

\begin{CodeInput}
R> mycache <- buildscorecache(data.df = asia, data.dist = distrib, 
+      max.parents = 4, dag.retained = ~LungCancer | Smoking)
R> mp.dag <- mostprobable(score.cache = mycache)
R> fabn <- fitabn(object = mp.dag, create.graph = TRUE)
\end{CodeInput}
\end{CodeChunk}

The function \code{fitabn()} computes the parameter estimates from a
given structure, a list of distributions and a dataset. The asia dataset
is made of Bernoulli variable and then the parameter estimates are log
odds ratio. The method used is a Bayesian approach, thus the estimates
are the modes of the posterior distributions.

\begin{CodeChunk}

\begin{CodeInput}
R> fabn
\end{CodeInput}

\begin{CodeOutput}
The ABN model was fitted using a Bayesian approach. The estimated modes are:

$Asia
Asia|(Intercept) 
           -4.77 

$Smoking
Smoking|(Intercept) 
              0.012 

$Tuberculosis
Tuberculosis|(Intercept) 
                   -4.72 

$LungCancer
LungCancer|(Intercept)     LungCancer|Smoking 
                 -4.28                   2.26 

$Bronchitis
Bronchitis|(Intercept)     Bronchitis|Smoking 
                 -0.85                   1.78 

$Either
 Either|(Intercept) Either|Tuberculosis   Either|LungCancer 
              -11.4                19.1                21.1 

$XRay
XRay|(Intercept)      XRay|Either 
           -3.09             8.30 

$Dyspnea
Dyspnea|(Intercept)  Dyspnea|Bronchitis      Dyspnea|Either 
              -2.08                3.31                2.19 

Number of nodes in the network: 8 .
\end{CodeOutput}
\end{CodeChunk}

There is an arrow between the nodes Lung Cancer and Smoking. The
function \code{fitabn()} returns an intercepts which could be
interpreted as noise and an odds ratio of 2.26 on the logit scale. In
exponentiating it we get the classical odds ratio (OR: 9.58) which is a
measure of association between the two random variable. The OR is larger
than 1 meaning that Lung Cancer and Smoking are positivelly associated.

In this simple example, we have plenty of observations to estimate 15
parameters. In more realistic cases, we have to control for
over-fitting. This will be further discussed in
Section~\ref{subsec:overfitting}.

\hypertarget{subsec:alt}{%
\subsection{Alternative R packages}\label{subsec:alt}}

Not too surprisingly, there are several \proglang{R} packages available
for BN modelling, which often cover a particular model and data niche.
The \proglang{R} packages can be divided into two broad classes: the
ones targeting parameters and structure learning and the ones targeting
inference in BN models. Following \proglang{R} packages target
parameters and structure learning. The most popular \proglang{R} package
is the \proglang{R} package \pkg{bnlearn} \citep{Scutari:2010aa}. It has
implementations of most of BD scores but also information theoretic
scores for continuous and discrete mixed variables. Additionally, it has
implementations of multiple network structure learning via multiple
constraint-based and score-based algorithms. This is the most versatile
\proglang{R} package for BN modeling and should be the prefered primary
software choice. When focusing on a causal framework, the \proglang{R}
package \pkg{pcalg} is very popular \citep{Kalisch:2012aa}. It contains
an implementation of the PC-Algorithm that selects one class
representative of the network skeleton. When dealing with discrete
Bayesian networks only, the \proglang{R} package \pkg{catnet} allows
parameter and structure learning using likelihood-based criteria. The
\proglang{R} package \pkg{deal} \citep{bottcher2003deal} enables
learning BN with a mixture of continuous and/or discrete variables under
the conditional Gaussian distribution (restriction of discrete nodes
being only parent of continuous ones). Other useful \proglang{R}
packages are actively maintained on CRAN, but none of them implement
scoring procedures that deal simultaneously with multiple exponential
family representatives in a Bayesian or likelihood-based framework and
allow for grouping correction and epidemiological adjustment. When
targeting inference the most used \proglang{R} packages are the
\pkg{gRbase} \citep{dethlefsen2005common} and \pkg{gRain}
\citep{hojsgaard2012graphical}. Indeed, those \proglang{R} packages do
not have any structural learning algorithm, thus the BN models should be
fully provided by the user. But they allows to manipulate efficiently
the models parameter to perform prediction and inference. The
\proglang{R} package \pkg{abn} is the only library for BN modelling
inference based on a fully Bayesian formulation. It contains a unique
implementation of an exact search algorithm. It targets systems
epidemiology applied research in providing to the user extensive outputs
easing biological findings interpretation and reporting. The Task View:
gRaphical Models in R (\url{CRAN.R-project.org/view=gR}) gives a
comprehensive overview of the different computing libraries available on
CRAN.

Aside from \proglang{R}, multiple implementations in other computing
environments such as \proglang{Weka} \citep{bouckaert2008bayesian}
(which is accessible in \proglang{R} via the \proglang{R} package
\pkg{RWeka}, \citealp{RWeka}), \proglang{MATLAB} \citep{murphy2001bayes}
or open source \proglang{python} or \proglang{C++} implementation exist.
In the epidemiological community, the \proglang{R} implementations are
the most popular ones due to the simplicity of using \proglang{R}.

\hypertarget{sec:method}{%
\section{Methodological background}\label{sec:method}}

This section describes the theoretical foundations of ABN. First, we
present the BN modelling paradigm. Then we extend this framework to
describe the ABN methodology. Finally, we embed the ABN methodology into
a learning scheme.

\hypertarget{subsec:bn}{%
\subsection{Bayesian Network}\label{subsec:bn}}

The idea of studying observational data using using a BN is quite old
\citep{pearl1985bayesian}. Formally, a BN for a set of random variables
\({\textbf{\textit{X}}}= \{X_1,\dots, X_n\}\) is a directed acyclic
graph (DAG) where the nodes are the random variables and the directed
links are the statistical dependencies between the nodes. A graph \(G\)
is the union of two sets: the set of \textit{nodes} or \textit{vertices}
and the set of \textit{arcs} or \textit{directed links}, \textit{arrows}
or \textit{edges}. Thus: \(G = (\mathbf{V},\mathbf{E})\), where
\(\mathbf{V}\) is a finite set of \textit{vertices} and \(\mathbf{E}\)
is a finite set of \textit{edges}. An index node \(X_j\) is said to be
the parent of a node \(i\) if the edge set \(\mathbf{E}\) contains an
edge from \(j\) to \(i\). A set of parents for a node \(j\) is denoted
by \({\mathbf{Pa}_j}\). Conversely, one can easily define the set of
children \({\mathbf{Ch}_j}\) for a node \(j\).

In a BN, the factorization of the joint probability distribution
P({\textbf{\textit{X}}}) through a so-called set of local probability
distributions is given by the \textit{Markov property}, which implies
that an index node \(X_j\) is dependent solely on its set of parents
\({\mathbf{Pa}_j}\)

\begin{align} 
 P({\textbf{\textit{X}}})=\prod_{j=1}^{n}P(X_j\mid {\mathbf{Pa}_j}).   \label{e:fact} 
\end{align}

In equation (\ref{e:fact}), the total number of nodes is denoted by
\(n\). A BN model, \({\mathcal{B}}\), is the union of the structure
\({\mathcal{S}}\) and the model particularization
\({\bm{\theta}}_{\mathcal{B}}\):
\({\mathcal{B}}=({\mathcal{S}}, {\bm{\theta}}_{\mathcal{B}})\). The
edges represent both \textit{marginal} and
\textit{conditional dependencies}. Collectively, they define the
structure or network which encodes the conditional independence through
graphical separation. \citet{verma1988influence} shows that if two nodes
are d-separated by a set of nodes, then the random variables are
conditionally independent through the set of variables. This theorem is
the starting point of a class of BN learning algorithms called
constraint-based approaches. In a BN, each node \(X_j\), with parent set
\({\mathbf{Pa}_j}\), is parametrised by a local probability
distribution: \(P(X_j\mid {\mathbf{Pa}_j})\). The choice of the
parametrisation is the source of an alternative class of learning
algorithms called score based approaches.

The Markov blanket (MB) of a node is the set of parents, children and
co-parents, i.e., the parents of the specified children. An interesting
property of the MB is that this is the set of nodes that fully inform an
index node. For example, in the right network of Figure \ref{fig:asia},
the MB of index node \code{"Dyspnea"} consists of nodes
\code{"Bronchitis"} and \code{"Either"}, whereas the MB of index node
\code{"Either"} consists of nodes \code{"XRay"}, \code{"Dyspnea"},
\code{"Tuberculosis"}, \code{"LungCancer"} and \code{"Bronchitis"}.
Indeed, a node is conditional independent of any non-descending node in
a BN given its parents.

\newpage

\hypertarget{subsec:abn}{%
\subsection{Additive Bayesian Network formulation}\label{subsec:abn}}

An ABN model, \({\mathcal{A}}\), is a graphical model that extends the
usual generalized linear model (GLM) to multiple dependent variables
through the factorization of their joint probability distribution
\citep{lewis2013improving}. An \({\mathcal{A}}\) model assumes that each
node is a GLM where the covariates are the parents and the distribution
depends on the index node.

Given an index node \(X_j\), a set of parents \({\mathbf{Pa}_j}\), and
using the classical notation for the exponential family parametrization
\citep{pitman_1936}

\begin{align}
   P(X_j \mid {\mathbf{Pa}_j}) &= 
    \exp\bigl(\eta(\theta_j)  T({\mathbf{Pa}_j}) - A(\theta_j)\bigr) \mathsf{d} H({\mathbf{Pa}_j}), \label{eq:expofam} 
\end{align}

where the functions \(\eta\), \(T\), \(A\), \(H\) may be node-dependent
(the indices omitted to simplify the notation) and where the parameters
\(\theta_j\) incorporate the configuration of the parents' node. For
example, in the case of binary variables, i.e., \(X_j\in\{0,1\}\),
equation (\ref{eq:expofam}) is simplified when using the classical logit
link function to \begin{align}
   P(X_j \mid {\mathbf{Pa}_j}) &= \text{logit}^{-1}(\theta_j)=\text{expit}(\theta_j),
\end{align} resulting in the classical logistic regression models for
all the nodes.

The ABN modelling technique is situated in a small data analysis niche.
It targets a dataset composed of variables issued from a mixture of
different distributions. The main focus is that the final model should
be interpretable. An \({\mathcal{A}}\) is called additive in the sense
that the effect of the parents is assumed to be additive in the
exponential family link scale.

\hypertarget{subsec:learn}{%
\subsection{Learning algorithm}\label{subsec:learn}}

Many learning strategies have been proposed depending on the ultimate
modelling goal being in concordance with the research question. To
perform inference in BN, one needs a probabilistic model to compare the
networks and a search algorithm to select the optimal structure. If we
restrict ourselves to a purely non-dynamic network (either using
temporally dependent data or a dynamical network) and observational
data, we can propose two main strategies. The first is constraint-based
approaches, where one learns the BN using statistical independence
tests. The optimal network is identified using the reciprocal
relationship between graphical separation and conditional independence
\citep{spirtes2001anytime}. The second popular approach is based on
structural scoring, where each candidate network is scored and the one
which has the highest score is kept. In practice, this is
computationally intractable for the typical number of variables involved
in a research project. The number of possible networks is massive and
increases super-exponentially with the number of nodes
\citep{robinson1977counting}. A practical workaround is to use a
decomposable score, i.e., a score that is additive in terms of the
network's nodes and depends only on the parents of the index node. This
approach is very close to the classical model selection in statistics
\citep{zou2017model}.

The scoring approach paradigm requires that the scores represent how
well the structure fits the data \citep{zou2017model}. Many scores have
been proposed for discrete BNs in a Bayesian context
\citep{daly2011learning} that aims at maximizing the posterior
probability. Indeed, \citet{heckerman1995learning} propose the so-called
Bayesian Dirichlet (BD) family of scores which use a Dirichlet prior.
The BD family regroups the K2, BDeu, BDs and BDla scores
\citep{scutari2018dirichlet}. For continuous BN with multivariate
Gaussian data, using an inverse Wishart prior leads to the BGe score
\citep{consonni2012objective}. By analogy, the scores can be adapted to
a mixture of variables such as information theoretic scores. They are
less suitable from a theoretical perspective but more polyvalent. Scores
within a frequentist framework have been proposed
\citep{daly2011learning}, such as Bayesian Information Criterion (BIC),
Akaike Information Criterion (AIC) and Minimum Description Length (MDL)
\citep{daly2011learning}. They all have in common a goodness-of-fit part
and a penalty for model complexity. A direct and natural extension of
this idea, implemented in the \proglang{R} package \pkg{abn}, uses the
posterior score in a Bayesian regression settings. When applied to BN
scoring, it is called marginal likelihood \citep{mackay1992bayesian}, as
the likelihood is marginalized in the parameter space for the
estimation. In a Bayesian setting, the parameter prior acts as a penalty
term.

The model learning phase involves two parts: 1. Network, skeleton or
structure learning \({\mathcal{S}}\); and, 2. Parameter learning where
the model parameter is \({\bm{\theta}}_{\mathcal{B}}\). Hence, in a
Bayesian framework, constructing an ABN model \({\mathcal{A}}\) given a
set of data \(\mathcal{D}\)

\begin{align}
& P({\mathcal{A}}\mid \mathcal{D}) = \underbrace{\,P( \theta_{\mathcal{A}}, \mathcal{S}\mid\mathcal{D})\,}_{\text{model learning}} =
\underbrace{\,P(\theta_{\mathcal{A}}\mid\mathcal{S},\mathcal{D})\,}_{\text{parameter learning}} ~ \cdot  \underbrace{\,P(\mathcal{S}\mid\mathcal{D})\,}_{\text{structure learning}}.
\end{align}

The two learning steps are interconnected. Several efficient algorithms
have been proposed for both learning procedures. In order to learn the
relationships between variables, the conditional probability
distributions should be fitted. In a frequentist setting, this can be
done using the classical Iterative Reweighed Least Square (IRLS)
algorithm \citep{faraway2016extending}. The structure selection step can
be done using a heuristic or exact approach.

\bigskip

An interesting feature of the ABN methodology is to be able to impose
external expert knowledge. Indeed, in most of applied data analysis,
some part of the network is known. For example, if two random variables
are timely related the direction of the possible arrow is known. Or if,
based on existing literature a possible connection is known to be
expected. The \proglang{R} package \pkg{abn} allows such external causal
input through a banning or a retaining matrix. Those matrices are used
to compute the list of valid parent combination. However, a more
theoretically sounding approach, suggested by
\citet{heckerman1995learning}, is to augment the observed data with
synthetic data that represents the causal belief. The practical
feasibility in an ABN analysis remains an open question
\citep{mccormick2013using}.

\newpage

\hypertarget{sec:pkg}{%
\section{The R Package abn}\label{sec:pkg}}

This section describes the functionalities of the \proglang{R} package
\pkg{abn} in categorizing them compare to their final objectives. We
provide insights about how the functions are implemented. We show
simulations that compare the model parameter estimations using a
Bayesian or an MLE approach and how different scores perform to retrieve
networks.

The \proglang{R} package \pkg{abn} has three level of functionalities
for different purposes:

\begin{itemize}
\item \textbf{Core functions:} aiming at performing an ABN analysis (scoring and learning)
\item \textbf{Ancillary functions for analysis:} aiming at supporting the analysis by enhancing plotting abilities, accounting for the uncertainty in the structure through link strength estimation, and comparing structure or getting structural metrics
\item \textbf{Ancillary functions for simulation:} aiming at helping in simulating ABN models by simulating DAGs and ABN data.
\end{itemize}

A classical ABN analysis is the sequential application of three R
functions see Section \ref{sec:motivatingexample} (typically:
\code{buildscorecache()}, \code{fitabn()} and \code{mostprobable()}).
The reason why the analysis is divided into three functions is to let
more freedom to the end user to tune parameter and to have a better
control over possible learning/fitting issues.

\hypertarget{set-of-core-functions}{%
\subsection{Set of core functions}\label{set-of-core-functions}}

The two main core functions for scoring are \code{buildscorecache()} and
\code{fitabn()}. The former is essentially a wrapper of the latter.
Those functions have a Bayesian and an MLE implementation which are not
equivalent in their output (there is an argument \code{method} that can
be bayes or mle to choose the implementation). They require minimally a
named dataset, the named list of the nodes' distribution and an upper
limit for network complexity. \code{buildscorecache()} first computes an
empty list of valid parent combinations using banning and retaining
input matrices (which are assumed to be empty by default). Then it
iterates through the cache to score the candidate piece of network. In
the Bayesian implementation, \code{buildscorecache()} and
\code{fitabn()} estimate a Bayesian regression using the following
parameter priors: weekly informative Gaussian priors with mean 0 and
variance 1000 for each of the regression parameters of the model (both
binomial and Gaussian), as well as diffuse Gamma priors (with shape and
scale equal to 0.001) for the precision parameters in Gaussian nodes in
the model using an internal \proglang{INLA} code.

\code{buildscorecache()} within an MLE setting uses an IRLS algorithm
depending on the given list of distributions at each step of the scoring
process. For the special case of the binomial nodes, the usual logistic
regression is tried. If it fails to estimate the given problem, a
bias-reduced tailored algorithm is used (Firth's correction). It is know
to improve accuracy of regression coefficients in presence of separation
\citep{van2016no}. If, however, the algorithm fails to return a finite
estimate, some predictors are sequentially removed until the design
matrix becomes fully ranked. These three steps ensure the \proglang{R}
package \pkg{abn} to be able to score a dataset even if there is data
separation.

The \code{fitabn()} function score a given network. It requires a valid
DAG, a named dataset and a named list of distributions. It returns the
list of score for each node, the parameter estimates, the standard
deviation and the \(p\)-values. Special care should be taken when
interpreting and displaying the \(p\)-values. Indeed, the DAG has been
selected using goodness of fit metric, so at least adjustment methods
should be used.

A unique feature of \code{fitabn()} and \code{buildscorecache()} is the
possibility to take one-layer clustering into account. In some
situations, data collection has a clear grouping aspect. Therefore,
there is a potential risk for non-independence between data points from
the same group that could cause over-dispersion. This can lead to
analyses which are over-optimistic, as the true level of variation in
the data is under-estimated.

To account for this additional variability, a random effect is
introduced. Thus, each node becomes a Generalized Linear Mixed Model
(GLMM) \citep{breslow1993approximate, faraway2016extending} instead of a
GLM, but in a Bayesian setting. This implies introducing an additional
correlation structure via random effects, i.e., adjusting for correlated
residuals or non-constant variance. The rationale is that standard
errors are underestimated during the first search so the DAG produced
during the structure search have more rather than less structure. That
structure could then be trimmed during the MCMC part.

We therefore compute the posterior distribution and check if it widens.
In such as case, we must take the clustering in the scoring scheme used
into account. From an applied perspective, the major limitation is the
computational burden of this approach. Indeed, if the clustering is
unlikely to impact the estimates, it is preferable not to take it into
account. The model is then much simpler and parsimonious. The clustering
adjustment could be performed in a subset of the nodes.

For learning BN, two types of algorithms are implemented: exact and
heuristic searches. The exact procedure, \code{mostprobable()}, runs the
exact order-based structure discovery approach of
\citet{koivisto2004exact} to find the most probable network. Its input
is a cache of precomputed scores as from \code{buildscorecache()}, the
desired score and structural prior to use. As described in
\citet{koivisto2004exact}, \code{mostprobable()} uses dynamic
programming to marginalize orders analytically. It cannot handle more
than 25 nodes, but on similar size problem, it outperforms any MCMC and
probabilistic approaches. It is implemented with two structural priors:
the koivisto prior that states that different cardinalities of parents
are considered to be equally likely a priori; and a structurally
uninformative where parent combinations of all cardinalities are equally
likely. The heuristic search algorithms implemented are the
hill-climber, the Tabu and the simulated annealing in the function
\code{searchHeuristic()}. The function requires a cache of pre-computed
scores, the desired score and some method-dependent arguments.

\hypertarget{subsec:dag}{%
\subsection{Defining a DAG}\label{subsec:dag}}

To specify a DAG, the \proglang{R} package \pkg{abn} uses an adjacency
matrix, i.e., a named square matrix with an entry in the \(ij\)th entry
if there is an arc from \(j\) (parent) to \(i\) (child).

The \proglang{R} package \pkg{abn} also recognizes a formula-like
syntax, similar to the classical \proglang{R} functions \code{lm},
\code{glm}, etc. A typical formula is
\code{~ node1|parent1:parent2 + node2:node3|parent3}. The formula
statement has to start with \code{~}. In this example, \texttt{node1}
has two parents (\texttt{parent1} and \texttt{parent2}). \texttt{node2}
and \texttt{node3} have the same \texttt{parent3}. The parents names
have to exactly match those given in data.df. The symbol \texttt{:} is
the separator between either children or parents, the symbol
\texttt{\textbar{}} separates children (left side) and parents (right
side), the symbol \texttt{+} separates terms and \texttt{.} replaces all
the variables in \texttt{data.df}. Due to this feature, those symbols
cannot be used in the names of the random variables. The banned matrix
could be produced by the following statement \code{~ female|.}.

\hypertarget{ancillary-functions-for-analysis}{%
\subsection{Ancillary functions for
analysis}\label{ancillary-functions-for-analysis}}

\code{plotabn()} and \code{tographiz()} are useful functions for
plotting DAGs. \code{plotabn()} can display a DAG with fitted values and
arrows with thickness proportional to arc strength.

The concept of link strength for discrete BN was introduced in
\citet{boerlage1992link}. A good unpublished overview is given by
\citet{ebert2009tutorial}. This method is especially useful to account
for the uncertainty when using BN to model. Indeed, an arc is either
present or absent. This strong dichotomisation of structural relations
in BN is often hard to interpret. The link strength tends to give a
proxy for arc support by the data and in an applied perspective is
complementary to the regression coefficients. In practice, the function
\code{linkStrength()} discretizes the dataset using a large choice of
histogram rules and then computes multiple link strength metrics. Then,
the estimates are plugged in the definition of the entropy to return the
so-called empirical entropy. A well-known problem of empirical entropy
is that the estimations are biased due to the sampling noise. It is also
known that the bias decreases as the sample size increases. The mutual
information estimation is computed from the observed frequencies through
a plug-in estimator based on the entropy.

Two functions are useful for comparing DAGs. The \code{compareDag()}
function returns multiple graph metrics to compare two DAGs: the
confusion matrix, or error matrix in the machine learning literature.
\code{compareDag()} provides a list with the true positive rate, the
false positive rate, the accuracy, the G-measure, the F1-score, the
positive predictive value, the false omission rate and the
Hamming-Distance. The \code{infoDag()} returns a list for standard
metrics for describing a DAG that contains the number of nodes, the
number of arcs, the average Markov blanket size, the neighborhood
average set size, the parent average set size and children average set
size.

The \proglang{R} function \code{scoreContribution()} computes the score
contribution of each individual observation to the total network score
and additionally returns the diagonal entries of the hat matrix. This
function attempts to produce influential measures adapted to Bayesian
Networks models.

\hypertarget{ancillary-functions-for-simulation}{%
\subsection{Ancillary functions for
simulation}\label{ancillary-functions-for-simulation}}

To simulate DAGs and ABN data, the functions \code{simulateDag()} and
\code{simulateAbn()} are provided. The function \code{simulateDag()}
generates DAGs with an arbitrary arc density. To ensure acyclicity, it
samples a triangular adjacency matrix. The arc density is tuned with a
binomial sampling probability. The \code{simulateAbn()} generates ABN
data using the \proglang{R} package \pkg{rjags}
\citep{plummer2013rjags}. Simulating observations from a given structure
is done with a random number generator, respecting the node ordering
using \proglang{JAGS} \citep{plummer2003jags}. It first creates a BUG
file in the actual repository, then uses it to simulate the data. This
function produces a data frame. The purpose of those two functions is to
assess the effectiveness of the other functions of the \proglang{R}
package \pkg{abn}. But they could also be used to plan and conceive
systems epidemiology studies in assessing the necessary number of
samples in function of the expected effect size.

\hypertarget{sec:cs}{%
\section{Case study}\label{sec:cs}}

\begin{figure}[!ht]
  \centering
\includegraphics[width=\textwidth]{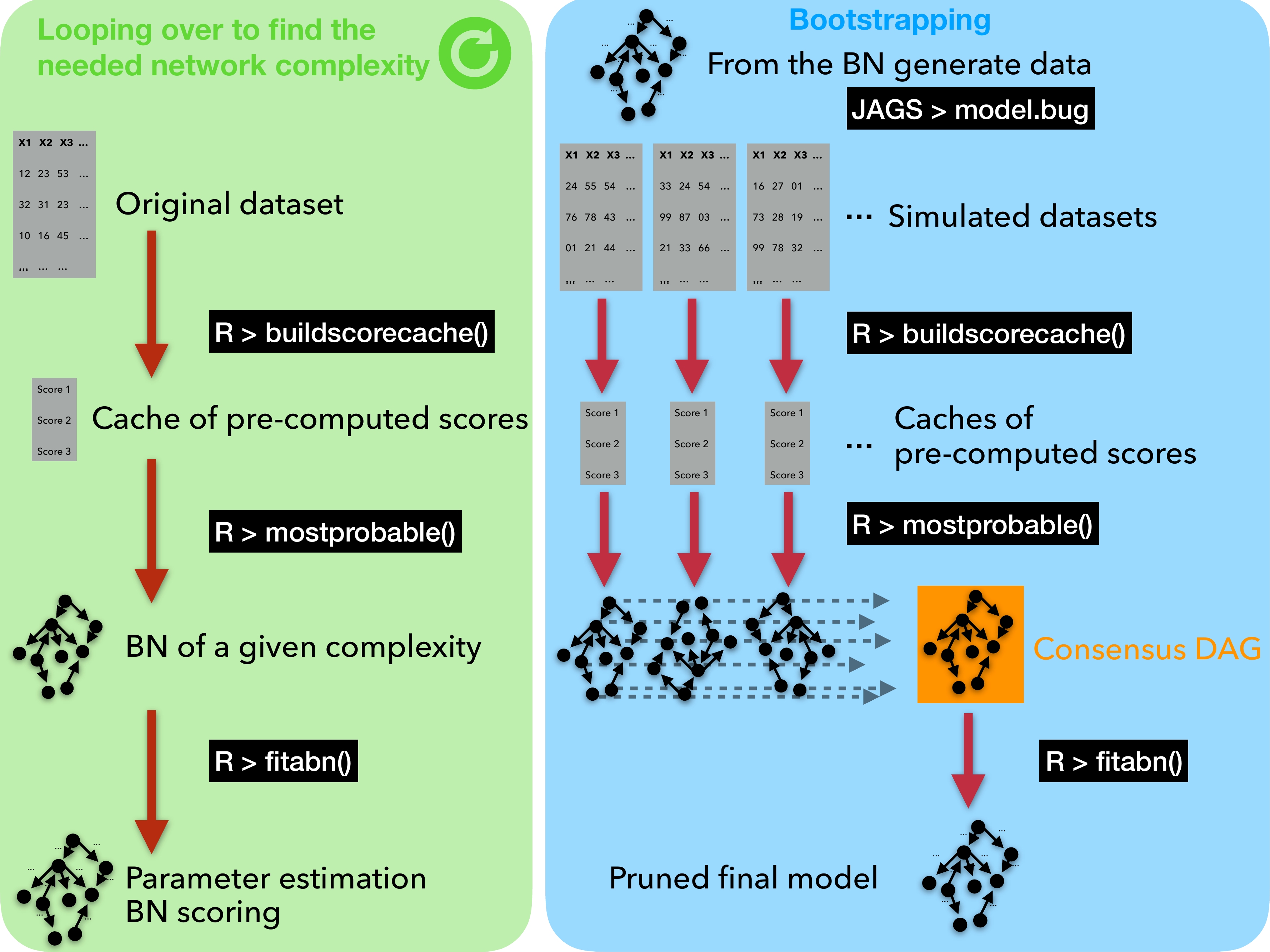}
  \caption{Schematic workflow diagram for a typical ABN analysis. In the {\color{green} green} square is the typical workflow for an ABN analysis. The {\color{blue} blue} square describes the bootstrapping procedure.}
  \label{fig:workflow}
\end{figure}

The purpose of this case study is to perform a fully reproducible and
transparent analysis of an open access observational dataset. The main
objective is to produce a reliable network together with the necessary
information to allow applied researchers to interpret and report it. The
final model is controlled for over-fitting, and some strategies for
controlling for clustering are presented. The code for the analysis is
provided to help disseminate the ABN approach in the systems
epidemiology community by addressing every step from an applied and
interpretative perspective. The proposed procedure for performing an ABN
analysis could be quite complex for a new user. Figure
\ref{fig:workflow} shows in \textcolor{green}{green} the workflow
diagram of a classical ABN analysis. It is the sequential application of
three functions: \code{buildscorecache()} for computing a cache of
pre-computed scores; a search algorithm such as \code{mostprobable()} or
\code{searchHeuristic()}; and \code{fitabn()} for fitting the final
model to the data. In \textcolor{blue}{blue}, the bootstrapping workflow
is presented. It is very similar to a classical ABN analysis, except
that it is based on simulated datasets to control for possible
over-fitting.

\hypertarget{data-description-and-library-loading}{%
\subsection{Data description and library
loading}\label{data-description-and-library-loading}}

The case study dataset collected in March 1987 is about growth
performance and abattoir findings in the commercial production of pigs
in 15 Canadian farms \citep{dohoo2003veterinary}. The data were
collected to study inter-relationships among respiratory diseases
(atrophic rhinitis and enzootic pneumonia), ascarid level and daily
weight gain. This dataset is well adapted to stress the unique ability
of BN modelling to disentangle complex relationships with observational
datasets. The data is an adapted version of the original dataset. It
consists of 341 observations of the 9 variables described in Table
\ref{tab:overview}. A nice feature of this dataset is that it is
composed of continuous, discrete and count distributed variables.
Furthermore, the dataset has a natural grouping feature due to the
different farms. This is used to showcase the capability of the
\proglang{R} package \pkg{abn} to control for a grouping effect.

\begin{table}[!ht]
\caption{\label{tab:overview}Description of the variables.}
\begin{tabular}[t]{lll}
\toprule
Variable & Meaning & Distribution\\
\midrule
AR & presence of atrophic rhinitis & Binomial\\
pneumS & presence of moderate to severe pneumonia & Binomial\\
female & sex of the pig (1=female, 0=castrated) & Binomial\\
livdam & presence of liver damage (parasite-induced white spots) & Binomial\\
eggs & presence of fecal/gastrointestinal nematode eggs & Binomial\\ 
& at time of slaughter & \\
wormCount & count of nematodes in small intestine at time of slaughter & Poisson\\
age & days elapsed from birth to slaughter (days) & Continuous\\
adg & average daily weight gain (grams) & Continuous\\
farm & farm ID & Discrete\\
\bottomrule
\end{tabular}
\end{table}

\begin{CodeChunk}
\begin{figure}[h]

{\centering \includegraphics{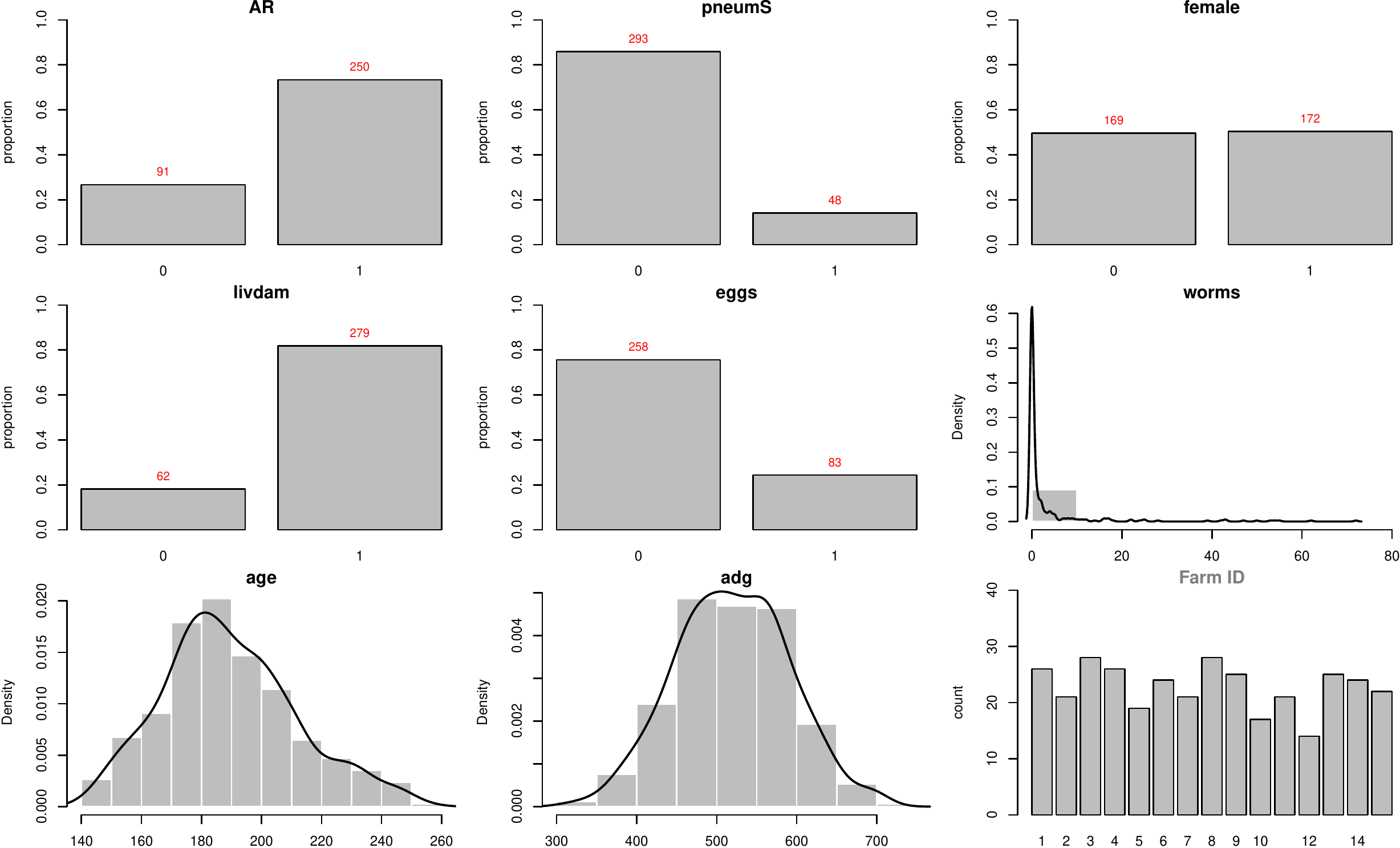} 

}

\caption{\label{fig:dist} Descriptive distributions of the variables.}\label{fig:dataDist}
\end{figure}
\end{CodeChunk}

The \proglang{R} package \pkg{abn} and its dependencies are available
from CRAN. To fully profit from the case study presented below, other
software needs to be installed: Rgraphviz \citep{hansen2016rgraphviz}
from bioconductor; or alternatively the \proglang{R} package
\pkg{DiagrammeR} \citep{diagrammer} and the \proglang{JAGS} computing
library \citep{plummer2003jags}. One can load the data using the
following code

\begin{CodeChunk}

\begin{CodeInput}
R> data("adg", package = "abn")
\end{CodeInput}
\end{CodeChunk}

\hypertarget{data-preparation-and-package-loading}{%
\subsection{Data Preparation and package
loading}\label{data-preparation-and-package-loading}}

Each variable in the model needs to be associated with a distribution.
Thus, one needs to create a named list that contains all the variables
and the corresponding distributions. Currently, the available data
distributions are binomial, gaussian, poisson and multinomial, where the
last distribution is available for MLE fitting only. The data of this
example is well defined, but in general one needs to coerce binary or
multinomial variables to factors

\begin{CodeChunk}

\begin{CodeInput}
R> dist <- list(AR = "binomial", pneumS = "binomial", 
+      female = "binomial", livdam = "binomial", eggs = "binomial", 
+      wormCount = "poisson", age = "gaussian", adg = "gaussian")
\end{CodeInput}
\end{CodeChunk}

A nice feature of the \proglang{R} package \pkg{abn} is the ability to
input prior information about structural beliefs about the data to guide
the structure search. Indeed, prior expert knowledge is common in
systems epidemiology. In this case study, it is reasonable to assume
that none of the variables in the model are going to affect the sex of
the animal, an inborn trait. To encode this information, we ban all the
arcs going to \code{female} by setting their value to 1 (banned),
opposite to the default value 0 (no banned) in an adjacency matrix-like
formulation. The rows are children and the columns are parents of the
index nodes.

\begin{CodeChunk}

\begin{CodeInput}
R> print(banned)
\end{CodeInput}

\begin{CodeOutput}
          AR pneumS female livdam eggs wormCount age adg
AR         0      0      0      0    0         0   0   0
pneumS     0      0      0      0    0         0   0   0
female     1      1      0      1    1         1   1   1
livdam     0      0      0      0    0         0   0   0
eggs       0      0      0      0    0         0   0   0
wormCount  0      0      0      0    0         0   0   0
age        0      0      0      0    0         0   0   0
adg        0      0      0      0    0         0   0   0
\end{CodeOutput}
\end{CodeChunk}

By default, the \proglang{R} package \pkg{abn} assumes no banned nor
retained arcs. See Section~\ref{subsec:dag} on how to specify banned or
retained arcs by using a formula-like syntax.

\newpage

\hypertarget{structure-search}{%
\subsection{Structure search}\label{structure-search}}

For computational reasons, it is advised to constrain the maximum number
of parents allowed for each node. We start to compute a cache of
pre-computed scores with a single parent per node and increase
subsequently the number of parents until the network score of the
optimal structure does not get larger, even when more parents are
allowed. Based on the cache of pre-computed scores, an exact search
using the function \code{mostprobable()} is performed and the network
score is computed. In \proglang{R} this is done using a simple
\code{for} loop with the functions \code{buildscorecache()},
\code{mostprobable()} and \code{fitabn()}.

\begin{CodeChunk}

\begin{CodeInput}
R> result <- vector("numeric", 7)
R> 
R> # ban: formula statement retain: not constrained
R> for (i in 1:7) {
+      mycache <- buildscorecache(data.df = as.data.frame(abndata), 
+          data.dists = dist, dag.banned = ~female | ., 
+          dag.retained = NULL, max.parents = i, method = "bayes")
+      mydag <- mostprobable(score.cache = mycache)
+      result[i] <- fitabn(object = mydag)$mlik
+ }
\end{CodeInput}
\end{CodeChunk}

Figure \ref{fig:nbparents} displays the network score as a function of
the number of parents. The maximum log marginal likelihood (--2709.25)
is achieved with 4 parents. As one can see, the network score increases
quickly at the beginning but then plateaus. A strong assumption is to
say that if increasing the network complexity by one does not change the
network score, then the maximum was achieved. Indeed, nothing prevent
the maximum network score being constant when increasing the number of
parent but still changing further when increasing again. An alternative
is to run an analysis with a pre-defined network complexity. Another
approach is to use a heuristic search provided by
\code{searchHeuristic()}. The heuristic approaches, contrary to the
exact search, do not guarantee the achievement of the optimal network
score. These approaches, however, do not suffer from computing
limitations, thus they are the only viable solution for large problems.
Many efficient heuristic search algorithms are implemented in
\pkg{bnlearn} \citep{Scutari:2010aa}.

\begin{CodeChunk}
\begin{figure}[h]

{\centering \includegraphics{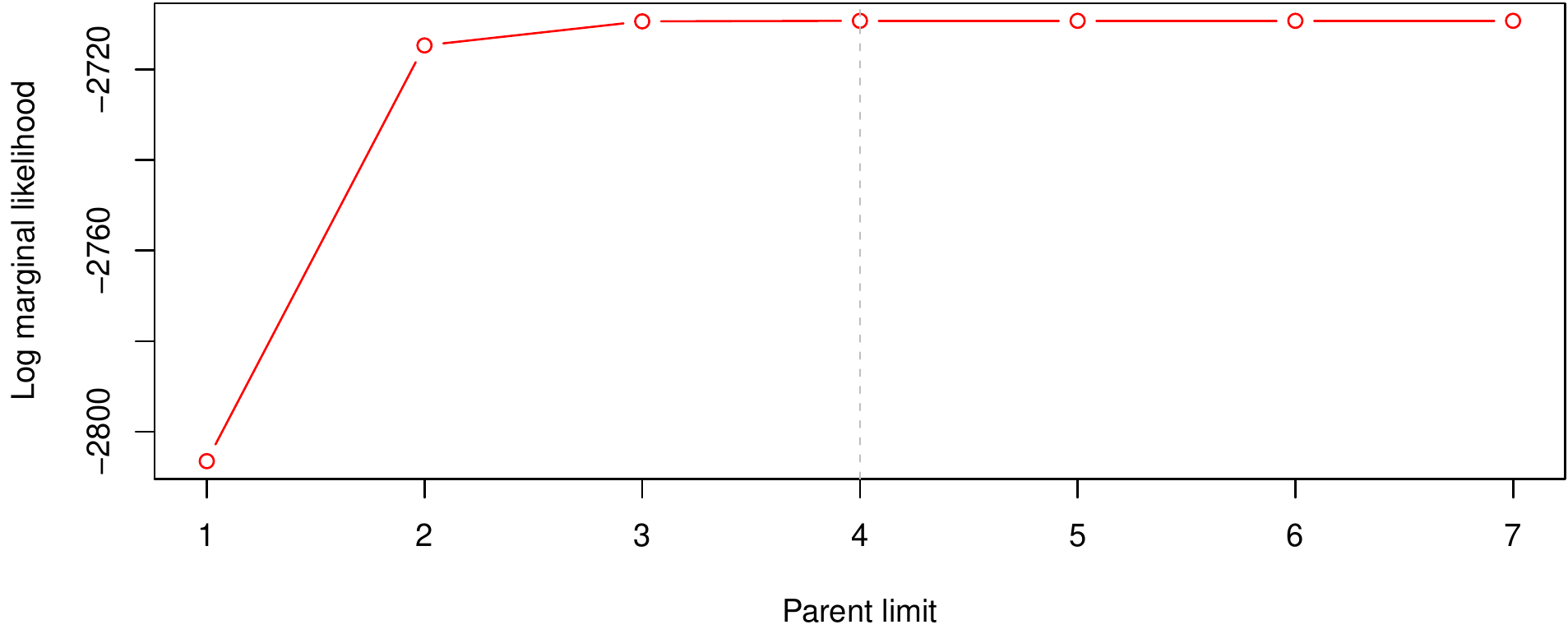} 

}

\caption{\label{fig:nbparents} Total network log marginal likelihood as a function of the number of parents.}\label{fig:mlik}
\end{figure}
\end{CodeChunk}

Figure \ref{fig:dag1} shows the DAG selected using \code{mostprobable()}
with a model complexity of 4 parents. This DAG has 10 arcs for 8
variables, so this is a relatively sparse model (the average number of
parent per node is 1.25, the average size of the Markov blanket is 3.75
and each node has on average 2.5 neighbors; as returned by function
\code{infoDag()} or \code{summary()}). The square nodes are Bernoulli
distributed, the oval nodes are normally distributed and the diamond
node is Poisson distributed.

\begin{CodeChunk}
\begin{figure}[h]

{\centering \includegraphics[width=0.4\linewidth]{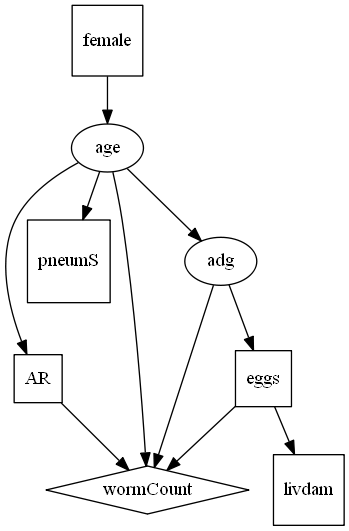} 

}

\caption{\label{fig:dag1} DAG selected using an exact search with a model complexity of four parents.}\label{fig:DAGshow}
\end{figure}
\end{CodeChunk}

\hypertarget{subsec:overfitting}{%
\subsection{Control for over-fitting}\label{subsec:overfitting}}

One major concern in BN modelling is the tendency for over-fitting.
Indeed, the number of observations is generally very limited in
comparison to the number of parameters in the model. The number of
possible models, i.e., DAGs, increases super-exponentially with the
number of random variables \citep{robinson1973counting}. The possible
number of DAGs with 25 nodes are about a googol (10\(^{100}\)).

More conceptually, the question remains of how to present the result of
a BN modelling analysis. Multiple strategies have been proposed to
create a summary network. One could perform multiple heuristic searches.
One common and very simple approach is to produce a single robust BN
model of the data mimicking the approach used in phylogenetics to create
majority consensus trees. A majority consensus DAG is constructed from
all the arcs present in at least 50\% of the locally optimal DAGs found
in the search heuristics. This creates a single summary network. The
function \code{searchHillclimber()} performs such an analysis. Rather
than using the majority consensus network as the most appropriate model
for data, an alternative approach would be to choose the single best
model found during a large number of heuristic searches using
\code{searchHeuristic()}, or if possible an exact search using
\code{mostprobable()}.

The sensible step of this approach is to set the necessary number of
searches needed to be run to provide reasonable coverage of all the
features of the model landscape. Then this model should be adjusted for
possible over-fitting. As with the majority consensus network, which
effectivelly averages over many different competing models and therefore
should generally comprise only robust structural features. Choosing the
DAG from a single model search is far more likely to contain some
spurious features, especially when dealing with small datasets of around
several hundred observations. It is extremely likely to over-fit, as one
can easily demonstrate using simulated data.

A simple assessment of over-fitting can be made by comparing the number
of arcs in the majority consensus network with the number of arcs in the
best-fitting model found using an exact search or a large number of
heuristic searches. We have found that in larger datasets, the majority
consensus and best-fitting model can be almost identical, while in
smaller datasets the best-fitting models may have many more arcs,
suggesting a degree of over-fitting.

An advantage of choosing a DAG from an individual search is that, unlike
averaging over lots of different structures as in the construction of a
majority consensus network, the model chosen here has a structure which
was actually found during a search across the model landscape. In
contrast, the majority consensus network is a derived model which may
never have been chosen during even an exhaustive search. Indeed, it may
comprise contradictory features, a usual risk when averaging across
different models. In addition, a majority consensus network need also
not to be acyclic, although in practice this can be easily corrected by
reversing one or more arcs to produce an appropriate DAG. A simple
compromise between the risk of over-fitting when choosing the single
highest-scoring DAG and the risk of inappropriately averaging across
different distinct data generating processes is to prune the
highest-scoring DAG using the majority consensus model. In short,
element-wise matrix multiplication of the highest-scoring DAG and the
majority consensus DAG gives a new DAG which only contains the
structural features in both models. An alternative approach for tackling
this problem, would be to use Bayesian model averaging. Structural MCMC
will be discussed in Section~\ref{sec:summary}.

\citet{friedman1999data} presented a general approach for using
parametric and non-parametric bootstrapping to select BN models/DAG
structures. They showed that a non-parametric approach seems to converge
less rapidly in terms of number of samples but requires fewer
assumptions. The non-parametric bootstrapping approach is relatively
easy to implement and efficient in extracting robust features from the
data. The parametric approach can be implemented by using readily
available Markov chain Monte Carlo sampling software such as
\proglang{JAGS} or \proglang{WinBUGS}. The basic idea is to take the
structure with the best network score, code it, and use these samplers
to generate bootstrap datasets from this model. That is, independent
realizations from the model which can be used to generate datasets of
the same size as the observed data. Then the model search is repeated
treating the bootstrap data as the observed data. Generating many
bootstrap datasets and conducting searches on each allows us to estimate
the percentage support for each arc in the highest-scoring model. In
other words, we find out how many of the arcs in the highest-scoring
model can be recovered from a dataset of the size that was actually
observed. Obviously, the more data, the more statistical power and
recoverable structural features. Arcs with a lower level of support,
e.g., \textless{}50\%, can be pruned from the best fitting model,
assuming that these are potentially a result of over-fitting. The
resulting model - possibly with arcs pruned from the original model - is
the chosen model for the data. The 50\% threshold is arbitrary and could
possibly depend on the expected arc density or model complexity. Other
options for trimming or pruning arcs exist, but without further
information it seems to be a reasonnable uninformative guess. This
strategy is commonly used in phylogenetic trees.

While parametric bootstrapping as a general technique is well
established and conceptually elegant, it may in practice not be
computationally feasible. Even when taking the least demanding approach
of conducting only one heuristic search per bootstrap dataset, the
number of datasets/searches required to get robust support values for
each arc in the best fitting model may be large and beyond what is
reasonably possible even using high performance computing (HPC)
hardware.

\hypertarget{parametric-bootstrapping}{%
\subsection{Parametric bootstrapping}\label{parametric-bootstrapping}}

The chosen dataset allows us to do fast bootstrapping. Given a BN model
- a DAG structure plus parameter priors - the first step is to estimate
the posterior parameters. The second step is to implement the DAG
structure together with the posterior parameters in a BUG file. To make
the implementation as general as possible, we present an approach based
on \code{dcat()} from \proglang{JAGS} that discretizes each posterior
density across a fine grid.

The function \code{fitabn()} with \code{compute.fixed = TRUE} uses
Laplace approximations to estimate the posterior density of each
parameter in a BN model. An appropriate domain (range) for each
parameter must be supplied by the user using a bit of trial and error.
It is crucially important to give a sufficiently wide range so that all
of the upper and lower tails of the distribution are included, e.g., the
range should encompass where a density plot first drops to approximately
zero at each tail.

\begin{CodeChunk}
\begin{figure}[h]

{\centering \includegraphics{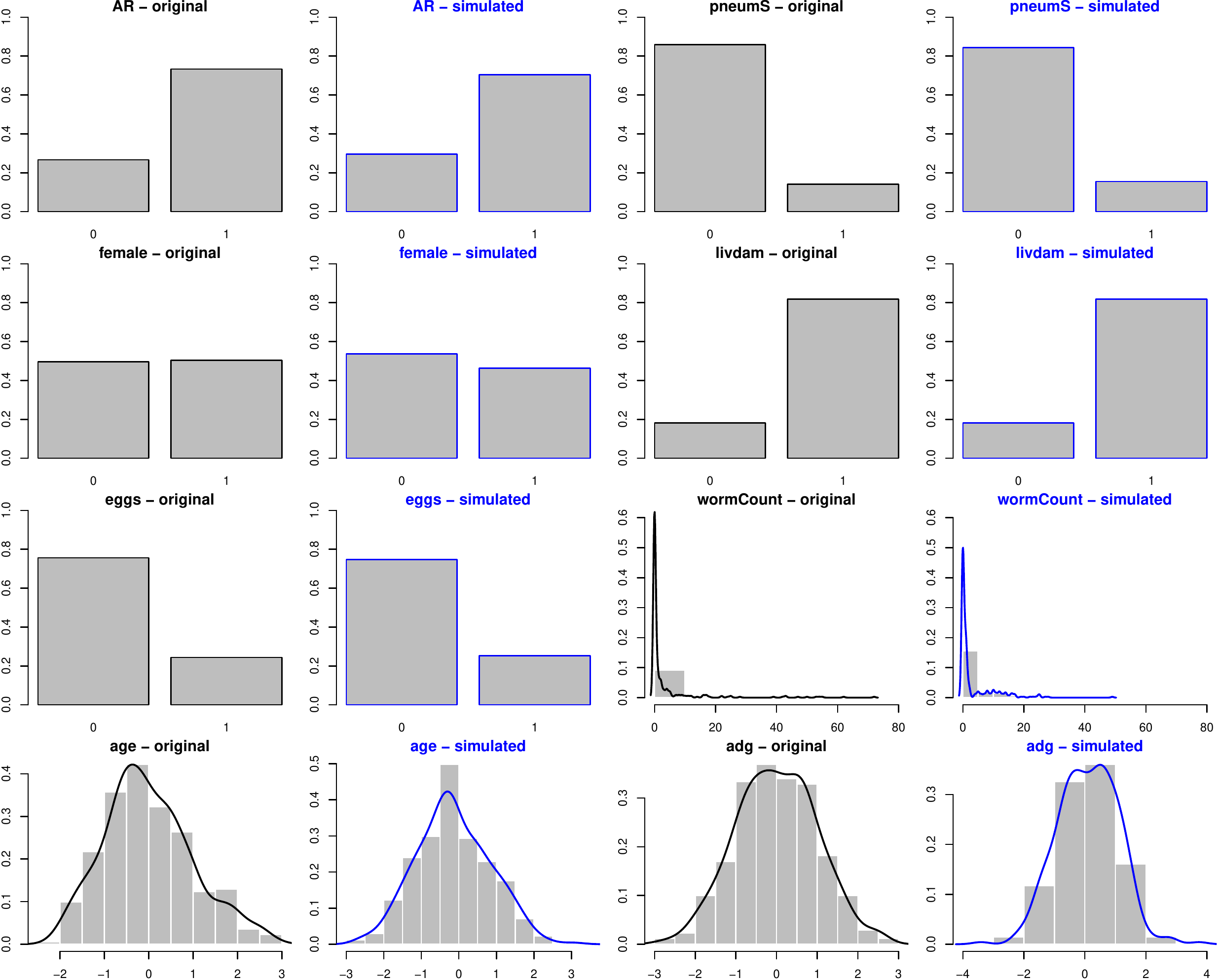} 

}

\caption{\label{fig:post} Example of simulated data (in blue) versus original data (in black).}\label{fig:testJAGS2}
\end{figure}
\end{CodeChunk}

Figure \ref{fig:post} compares one round of simulated data (in blue)
with the original data (in black). By default, \code{fitabn()} center
and standardizes continuous data. As one can see, the simulated data
looks fairly good except for the count data, which looks sub-optimally
simulated as the long right tail is under-represented in the simulated
data.

The code displayed in Annex \ref{sec:paramboot} performs 5000
bootstrapping steps from a model defined in a BUG file called
\code{model8vPois.bug} using the \proglang{R} package \pkg{rjags}. A
thinning of 20 is used to reduce auto-correlation in the simulated
samples. The global search is depicted with the pseudo-code given in
Algorithm \ref{alg:pseudoboot}.

\bigskip

\begin{algorithm}[ht]
\bigskip
\DontPrintSemicolon
\SetAlgoLined
\KwResult{A list of DAGs and bootstrapped dataset}
\SetKwInOut{Input}{Input}\SetKwInOut{Output}{Output}\SetKwInOut{Result}{Result}
\Input{A fitted ABN model, an original dataset, model complexity limit, a random seed, a number of desired bootstrap runs, a BUG file model, a thinning parameter}
\Output{List of matrix and dataframe}
\BlankLine
\For{1:n}{
    Read a BUG file and run JAGS over it\;
    Extract simulated samples\;
    Format data\;
    Compute a cache of pre-computed scores\;
    Run an exact search\;
    Fit an ABN model to the selected BN\;
}

\caption{Pseudo-code for bootstrapping ABN models.}\label{alg:pseudoboot}
\end{algorithm}

\bigskip

In the following text, the list of DAGs selected from the simulated
datasets is called the bootstrapped samples. In order to simulate the
variables of the dataset, we need to provide a model for each of them
using the aforementioned parameters estimates. For instance, the
binomial node \texttt{AR} in our DAG has one incoming arc coming from
the node \texttt{age}. In a regression setting this would be translated
into:\\
\begin{align} 
\text{logit}(\mathtt{AR}) = \alpha + \beta \cdot \mathtt{age} + \varepsilon,
\end{align} where \(\alpha\) is the intercept, \(\beta\) is the
regression coefficient for variable \texttt{age}, and \(\varepsilon\) is
the error term modeled by a binomial distribution. Then, the values of
\(\alpha\) and beta is sampled using the function \code{dcat()} within
\proglang{JAGS} from the original discrete distribution of parameters
generated by \code{fitabn()}.

Figure \ref{fig:table_arc} displays the arc distribution from the
selected BN from the simulated datasets, i.e., the bootstrapped samples.
The network selected from the original dataset had 10 arcs. This is an
indication of over-fitting from the original model.

\begin{CodeChunk}
\begin{figure}[h]

{\centering \includegraphics{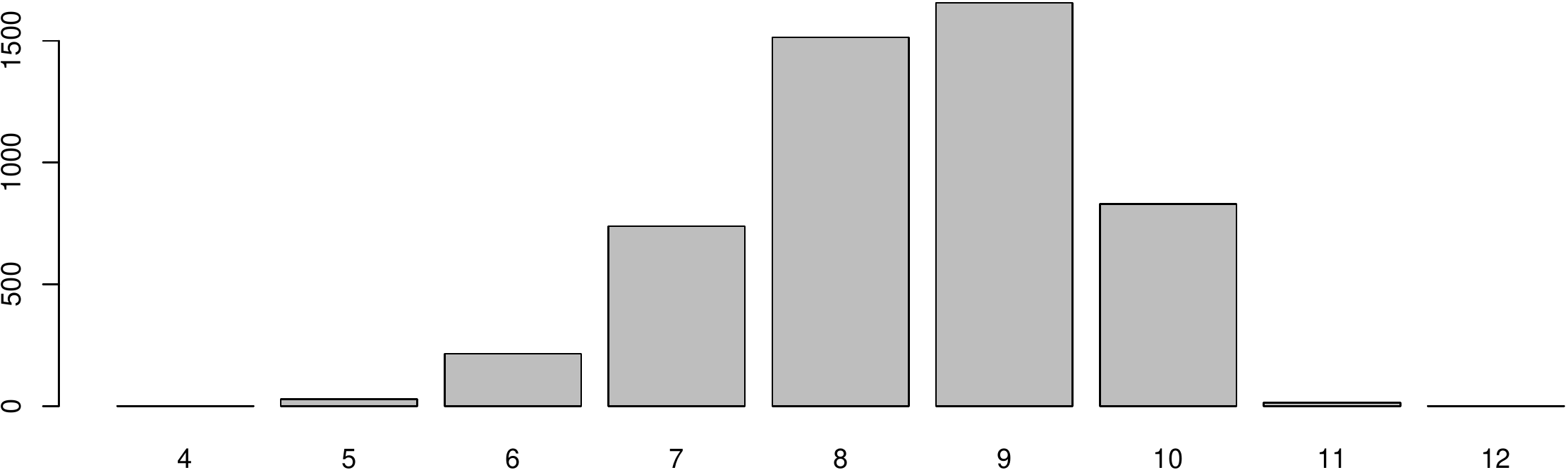} 

}

\caption{\label{fig:table_arc} Histogram of the distribution of the number of arcs in the bootstrapped searches}\label{fig:checkBoot}
\end{figure}
\end{CodeChunk}

In order to produce the final pruned DAG, we measure the prevalence of
each arc in the generated structures and retain only arcs present in at
least 50\% of the bootstrapped samples.

The matrix displayed below shows the percentage of arcs retrieved within
the bootstrap samples. As one can see, some arcs are clearly supported
or excluded, but some are very close to the 50\% threshold. This is not
surprising given the typical number of observations of an
epidemiological study (a few hundred). An alternative approach would be
to consider an undirected graph that includes all arcs supported
regardeless to direction in more than 50\% of the bootstrap samples.
This is justifiable due to the fact that the score used is approximately
the score equivalent. Thus, generally speaking, the data and ABN
methodology cannot discriminate between different arc directions;
therefore, considering arcs recovered in only one direction may be
overly conservative. This decision is likely to be problem-specific.

\begin{CodeChunk}

\begin{CodeInput}
R> print(perdag * 100, digits = 0)
\end{CodeInput}

\begin{CodeOutput}
          AR pneumS female livdam eggs wormCount age adg
AR         0      0      1      0    0         0  76   5
pneumS     1      0      0      0    0         0  40   8
female     0      0      0      0    0         0   0   0
livdam     0      0      0      0   53         2   0   0
eggs       0      0      0     25    0         0   8  75
wormCount 71      1      1      0  100         0 100  60
age       18     17     56      0    3         0   0  26
adg        1      2     11      0    6         0  74   0
\end{CodeOutput}
\end{CodeChunk}

\hypertarget{control-for-clustering}{%
\subsection{Control for clustering}\label{control-for-clustering}}

In the \code{adg} dataset, the data were collected in different farms
(random variable encoded as \code{farm}). A workaround was needed to
introduce additional correlation structure via random effects when
fitting the data to the structure to avoid overly optimistic parameter
estimates. Three strategies for clustering adjustment are possible in an
ABN analysis. They are ranked in terms of computational complexity:

\begin{itemize}
\item \textbf{Adjustment at the regression phase:} given the structure selected without adjustement (eventually controlled for overfitting), it is possible using the function \code{fitabn()} to adjust for clustering the regression coefficients.
\item \textbf{Adjustment at the bootstrapping phase:} given the structure selected without adjustement, one can alterate the code aiming at controlling for overfitting in adding a gaussian random effect in the model.
\item \textbf{Adjustement at the learning phase:} learning the structure with cluster adjusted scores with the function \code{buildscorecache()}. 
\end{itemize}

\hypertarget{adjustment-at-the-regression-phase}{%
\subsubsection{Adjustment at the regression
phase}\label{adjustment-at-the-regression-phase}}

The function \code{fitabn()} has the possibility to control internally
for clustering using the arguments group.var and cor.var. One can choose
which variable should be adjusted. On a personal computer, computing the
adjusted model takes 15 minutes to compute, whereas the unadjusted model
ran in less than a second. The grouping variable has 15 levels. In the
function \code{fitabn()}, we apply the grouping adjustment to all random
variables.

The following code will produce adjusted regression coefficients:

\begin{CodeChunk}

\begin{CodeInput}
R> marg.f.grouped <- fitabn(object = mydag, group.var = "farm", 
+      cor.vars = c("AR", "livdam", "eggs", "wormCount", "age", 
+          "adg"), compute.fixed = TRUE, n.grid = 1000)
\end{CodeInput}
\end{CodeChunk}

\hypertarget{adjustment-at-the-bootstrapping-phase}{%
\subsubsection{Adjustment at the bootstrapping
phase}\label{adjustment-at-the-bootstrapping-phase}}

In the BUG file, one can add a random effect for each level of the
clustering variable (\texttt{M}=15; the number of farms). The random
effect are typically chosen as normally distributed with mean zero and
precision parameter determined with a diffuse gamma prior. In practice
one modify the BUG file with a for loop over the number of level of the
clustering variable (\texttt{M}=15; the number of farms) for each
variable that should be adjusted. And a list of priors for precision
parameters. This approach aimes at verifying which of the selected arcs
are robust enough to pass the bootstrapping phase when correction is
applied to account for the additional variance produced by the
clustering. \newline

\begin{verbatim}
$> for(j in 1:M){\newline
$  % random effect for each group for variable adg\newline
$  rv.adg[j] ~ dnorm(0.0,prec.rv.adg);\newline
$}\newline
$> % priors definitions\newline
$> % rv priors\newline
$>\newline
$> prec.rv.adg ~ dgamma(1,5E-05);\newline
\end{verbatim}

\hypertarget{adjustment-at-the-learning-phase}{%
\subsubsection{Adjustment at the learning
phase}\label{adjustment-at-the-learning-phase}}

The finest adjustment is done directly at the learning phase when
pre-computing the network scores. This is also the most computationally
demanding and it can become numerically unstable given the number of
model that should be computed. This is done using the group.var and
cor.var arguments in \code{buildscorecache()}. In the following code the
same structural constrains have been used and the number of possible
parent is limited to 4.

\begin{CodeChunk}

\begin{CodeInput}
R> # recompute the cache of scores using GLMM
R> mycache <- buildscorecache(data.df = abndata, data.dists = dist, 
+      group.var = "farm", cor.vars = c("AR", "pneumS", 
+          "female", "livdam", "eggs", "wormCount", "age", 
+          "adg"), dag.banned = ~female | ., dag.retained = NULL, 
+      max.parents = 4)
R> 
R> # exact search
R> dag.adjusted <- mostprobable(score.cache = mycache)
R> 
R> # recompute the marginal using GLMM
R> marg.f.grouped <- fitabn(object = dag.adjusted, group.var = "farm", 
+      cor.vars = c("AR", "pneumS", "female", "eggs", 
+          "wormCount", "age", "adg"), compute.fixed = TRUE, 
+      n.grid = 100, control = list(max.mode.error = 0, 
+          epsabs.inner = 0.1, max.hessian.error = 0.5, 
+          epsabs = 0.1, error.verbose = FALSE, hessian.params = c(0.01, 
+              0.1), factor.brent = 10, loggam.inv.scale = 0.1))
\end{CodeInput}
\end{CodeChunk}

\begin{CodeChunk}
\begin{figure}[h]

{\centering \includegraphics{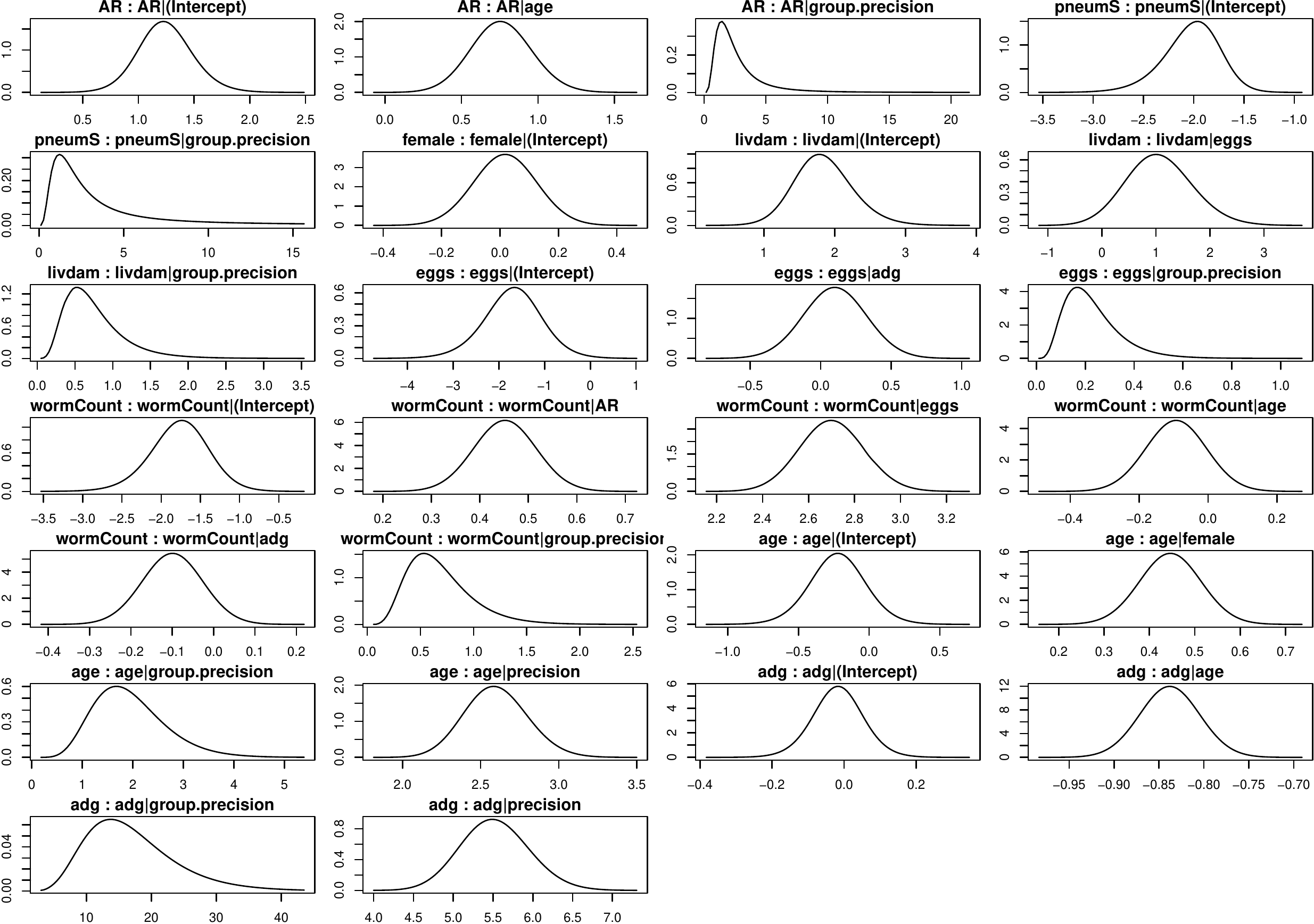} 

}

\caption{\label{fig:mp_plot_random} Marginal densities of model parameter corrected for random effect.}\label{fig:unnamed-chunk-11}
\end{figure}
\end{CodeChunk}

Figure \ref{fig:mp_plot_random} shows the parameter distribution
adjusted for clustering using random effect. The adjustement have been
applied to every nodes except the node female. After postproceesing, the
random effect applied to this node is poorly estimable. Indeed, the
number of female versus castrated pigs accross farms is the fairly
similar, which implies that the clustering has a negligeable effect. As
on can see, the quantiles have not widened much. Thus, we can use the
new parameter estimates.

\hypertarget{accounting-for-uncertainty-in-bn-models}{%
\subsection{Accounting for uncertainty in BN
models}\label{accounting-for-uncertainty-in-bn-models}}

Calculating the so-called link strength is useful for both visualization
and approximate inference, and it can be seen as a proxy for arc
uncertainty. The strength of the edges is a complementary metric to
regression coefficients. We use a link strength metric called the true
average link strength percentage (PLS), which expresses by how many
percentage points the uncertainty in variable \(Y\) is reduced by
knowing the state of its parent \(X\) if the states of all other parent
variables are known (averaged over the parent states using their actual
joint probability). The actual definition is:

\begin{align} 
\mathit{PLS}(X\to Y) = \frac{H(Y|Z)-H(Y|X,Z)}{H(Y|Z)}. \label{eq:ls} 
\end{align}

For the case of an arc going from node X to node Y and where the
remaining set of parents of Y is denoted as Z. H is the entropy. The
matrix below displays the percent link strength:

\begin{CodeChunk}

\begin{CodeInput}
R> print(LS, digits = 3)
\end{CodeInput}

\begin{CodeOutput}
             AR pneumS female livdam   eggs wormCount   age   adg
AR        0.000      0 0.0000      0 0.0000         0 0.089 0.000
pneumS    0.000      0 0.0000      0 0.0000         0 0.000 0.000
female    0.000      0 0.0000      0 0.0000         0 0.000 0.000
livdam    0.000      0 0.0000      0 0.0405         0 0.000 0.000
eggs      0.000      0 0.0000      0 0.0000         0 0.000 0.128
wormCount 0.282      0 0.0000      0 0.4068         0 0.434 0.376
age       0.000      0 0.0189      0 0.0000         0 0.000 0.000
adg       0.000      0 0.0000      0 0.0000         0 0.390 0.000
\end{CodeOutput}
\end{CodeChunk}

\hypertarget{presentation-of-the-results}{%
\subsection{Presentation of the
results}\label{presentation-of-the-results}}

The marginals represent estimates of the parameters at each node, i.e.,
the arcs in the DAG. As the variables are coming from different
distributions, they have different biological interpretations. In Figure
\ref{fig_final}, the square nodes are Bernoulli distributed, the oval
nodes are normally distributed and the diamond node is Poisson
distributed. The posterior marginals represent correlation coefficients
for Gaussian nodes, log rate ratios for Poisson nodes, and log odds
ratios for binomial nodes. Binomial and Poisson nodes need to be
exponentiated to get the odds ratios or rate ratios, respectively.

\begin{CodeChunk}
\begin{figure}[h]

{\centering \includegraphics[width=.4\linewidth]{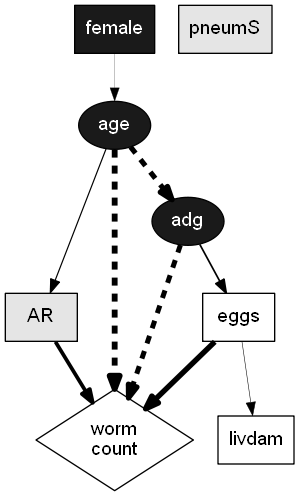} 

}

\caption[\label{fig_final}Final ABN model with arc width proportional to link strength]{\label{fig_final}Final ABN model with arc width proportional to link strength}\label{fig:fig_final}
\end{figure}
\end{CodeChunk}

\begin{table}[!h]
\caption{\label{tab:marg_post}Marginals posterior distribution of the parameter estimates}
\centering
\begin{tabular}[t]{lrrrrr}
\toprule
  & 2.5Q & median & 97.5Q & interpretation & LS\\
\midrule
AR|age & 1.46 & 2.14 & 3.19 & odds ratio & 0.09\\
livdam|eggs & 0.92 & 2.84 & 10.23 & odds ratio & 0.04\\
eggs|adg & 0.71 & 1.10 & 1.73 & odds ratio & 0.13\\
wormCount|AR & 1.38 & 1.57 & 1.79 & rate ratio & 0.28\\
wormCount|eggs & 11.38 & 14.85 & 19.59 & rate ratio & 0.41\\
wormCount|age & 0.77 & 0.91 & 1.08 & rate ratio & 0.43\\
wormCount|adg & 0.78 & 0.90 & 1.05 & rate ratio & 0.38\\
age|female & 0.31 & 0.44 & 0.58 & correlation & 0.02\\
adg|age & -0.90 & -0.84 & -0.77 & correlation & 0.39\\
\bottomrule
\end{tabular}
\end{table}

\newpage

\hypertarget{subsec:sim}{%
\section{Simulation study}\label{subsec:sim}}

Simulation studies are needed to test models and implemented methods. As
an illustration we show here one simulation that illustrate the
efficiency of the implemented scoring system. More simulation studies
are in the annexes: \ref{sec:dagsim} about DAGs structural metrics and
\ref{subsec:coefestim} about a quality assesment of the estimation of
the regression coefficients.

The parameters which are important from a simulation point of view are
the BN dimension, i.e., the number of nodes of the BN, the structure
density, i.e., average number of parents per node, and the sample size.
Additionally to those structure-wise metrics, important factors that
impact simulations are the intensity of the arc link, the variability of
the arc's distributions and the mixture of variables. Indeed, scores
used are only approximately Markov class-independent and the discrepency
increases when mixing distributions.

\hypertarget{subsec:scoreeff}{%
\subsection{Comparison of score efficiency}\label{subsec:scoreeff}}

In order to estimate the efficiency of the different scores implemented
in the \proglang{R} package \pkg{abn}, BNs with a given arc density have
been simulated, from which 20 datasets have been simulated with
different sample sizes. The metrics used to display performance of the
score are the true positive (number of arcs properly retrieved), false
positive (selecting an arc when there is not one) and false negative
(selecting no arc when there is one). The scores used are marginal
posterior likelihood in a Bayesian regression framework (abn), maximum
likelihood without penalization for complexity (mlik), AIC, BIC and MDL.

\begin{figure}[!ht] 
  \centering
\includegraphics[width=\textwidth]{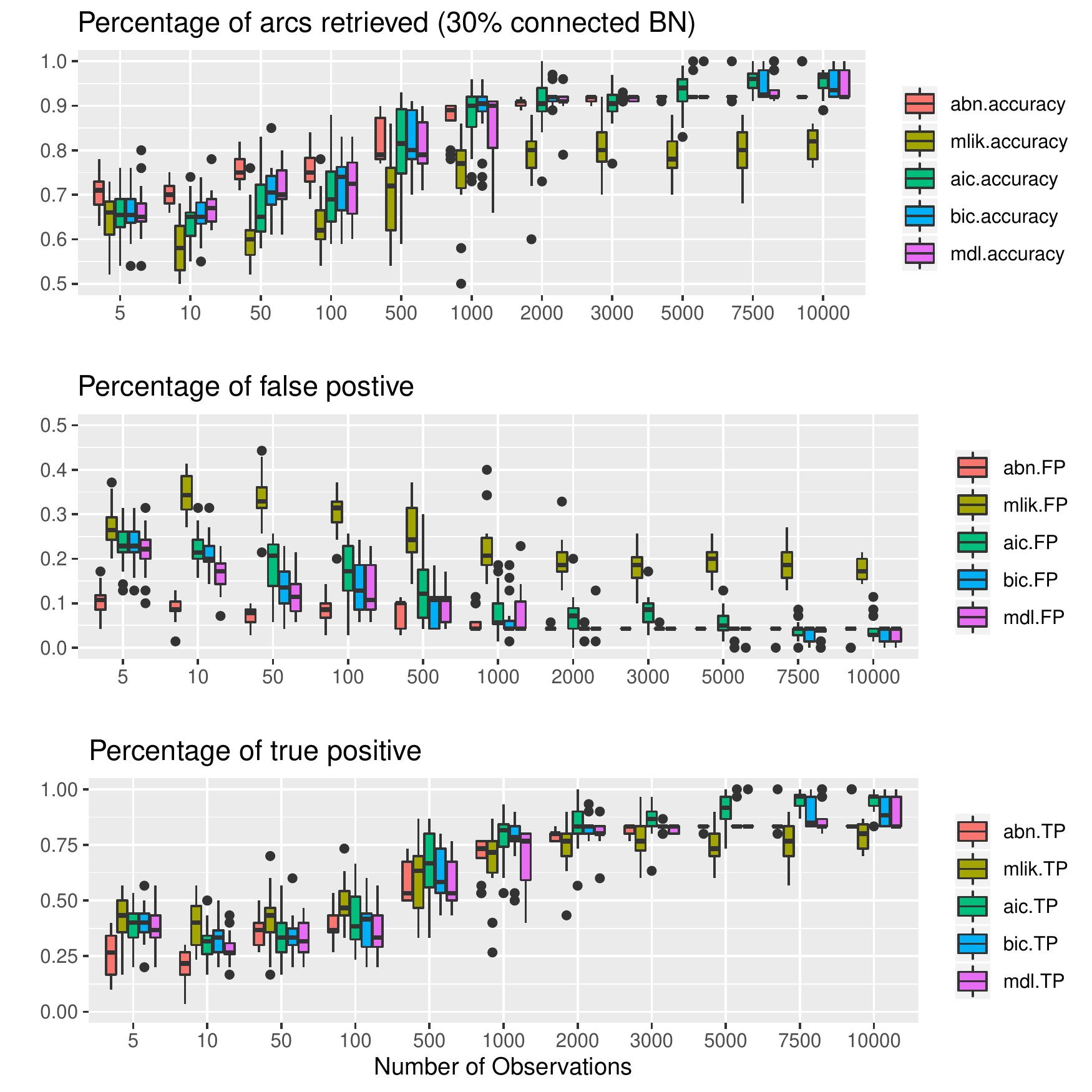}
  \caption{Comparison of score efficiency for marginal posterior likelihood (abn),  maximum likelihood (mlik), AIC, BIC and MDL for a given BN density in function of the number of observations using accuracy, true positive, false positive and false negative metrics.}
  \label{fig:score}
\end{figure}

Figure \ref{fig:score} shows for an increasing number of observation
\(n=5,10,50,\dots, 10000\) the efficience of five scores mentioned above
for a given BN density using boxplots to display the variability of the
distribution of different metrics. As one can see in Figure
\ref{fig:score}, the mlik (maximum likelihood) is a sub-optimal score
for BN learning as \citet{daly2011learning} expected. None of the other
scores are better or worse in general and they give different results
for a specific dataset. Interestingly from an applied perspective, the
marginal posterior likelihood seems to achieved very promising results
with a limited number of observations.

\hypertarget{sec:summary}{%
\section{Summary and discussion}\label{sec:summary}}

The \proglang{R} package \pkg{abn} is a free implementation of the ABN
methodology that allows users to fit Bayesian and MLE ABN models to
observational datasets. The \proglang{R} package \pkg{abn} contains
functions to analyze, select, plot and simulate ABN models. The
functions are designed to be usable with limited knowledge of
state-of-the-art Bayesian inference methods.

The \proglang{R} package \pkg{abn} suffers from several applied and
modelling limitations. Some of these limitations are inherent in the
Bayesian modeling framework some are tentatively tackled in an ABN
satellite suite of \proglang{R} packages. These packages are at
different development stages, but they are all designed to work
synergetically with the \proglang{R} package \pkg{abn} and hopefully
will be integrated in the long term.

From an applied perspective, researchers often have to extract a limited
number of variables from a large observational dataset. In an ABN
context, the exact search algorithm is computationally very sensitive to
the number of variables of the network. An ABN analysis targets
epidemiological modeling problems where a set of variables of importance
rather than a unique outcome can be clearly identified for the
modelling. In situations where no previous model exists and no clear
outcome could be identified to assess the predictive power of the
covariate, most of the classical model selection technique fails. Driven
by these findings, a variable selection approach that solves those
issues has been proposed to facilitate the ABN analysis. The
\proglang{R} package \pkg{varrank} \citep{varrank_gk, Kratzer:2018aa} is
an implementation of the minimum redundancy maximum relevance model
\citep{298224} that performs multi-outcome variable ranking. It can be
used prior to the \proglang{R} package \pkg{abn} to perform
dimensionality reduction and to focus on modelling important variables.
A possible alternative could be to use Random Forest's variable
importance that also supports multi-outcome formulation
\citep{strobl2007bias}.

A more theoretical limitation of the \proglang{R} package \pkg{abn} is
the fact that the parameter priors implemented are designed to be
non-informative. Bayesian model selection algorithms with uninformative
priors will asymptotically always select the simpler model, regardless
of the data. This is known as Lindley's paradox \citep{Lindley1957}. A
byproduct of more informative priors is the ability to guide the
posterior when the likelihood is poorly calculable. This could be useful
when there is data separation or in the case of paucity of data. Indeed,
while probably suitable in large datasets, flat priors can quickly turn
troublesome when the data is not informative for a parameter of interest
\citep{kratzer2019_BAYSM}. From a long-term perspective, it is certainly
of interest to equip \pkg{abn} with weakly informative priors whatever
the distribution to incorporate and extend the \proglang{R} package
\pkg{abn}.

One general limitation of the BN modelling approach and the \proglang{R}
package \pkg{abn} is the strong assumption of data independence coming
from the regression framework used. Multiple approaches have been
proposed to model temporal dynamics, such as Dynamic Bayesian Network
modelling (DBNs) \citep{murphy2002dynamic}, Temporal Nodes Bayesian
Networks (TNBNs) \citep{arroyo1999temporal}, VAR processes
\citep{ahelegbey2016bayesian} and state-space or hidden Markov models
\citep{le1999monitoring}.

The two pillars of the statistical inference are the estimation of the
effect size and the quantification of the model uncertainties. In a
regression context, this is achieved by reporting regression
coefficients and their confidence intervals (or poorly using solely the
coefficients' p-values). This is done in ABN analysis when reporting the
quantiles of the posterior distribution of the coefficients. However,
this is not fully satisfactory. Indeed, the uncertainties quantification
is performed conditionally to one model. A superior approach would be to
perform Bayesian model averaging, using MCMC over the possible
structures. The computed MCMC sample can be used to extract the arc
probability of presence or absence. In a broader perspective, it could
be possible to compute the probability of any structural query over the
MCMC sample. This approach is implemented in the \proglang{R} package
\pkg{BiDAG} \citep{bidag} or in the \proglang{R} package \pkg{mcmcabn}
\citep{kratzer2019single}. The former is based on an innovative order
search, whereas the latter is based on a fully structural formulation.
However, known to be slower in convergence and less efficient with large
networks than the \proglang{R} package \pkg{BiDAG}, the \proglang{R}
package \pkg{mcmcabn} has a fully transparent formulation regarding the
structural priors used. It eases the interpretation of findings.

\hypertarget{future-developments}{%
\section{Future developments}\label{future-developments}}

The case study showed that more exponential distributions should be
implemented in order to better grasp the diversity of variables of
interest in systems epidemiology. Indeed, zero-inflated variables are
poorly represented by the Poisson distribution. A negative binomial
would probably be a better alternative. This paper presents an analysis
that takes advantage of bagging to decrease model variance. Figure
\ref{fig:score} shows that information theoretic scores perform
differently depending on the context. Thus, there could be an
opportunity to construct a boosted score dedicated to the ABN
methodology. This idea should be addressed in a dedicated simulation
study.

\hypertarget{authors-contributions}{%
\section*{Authors contributions}\label{authors-contributions}}
\addcontentsline{toc}{section}{Authors contributions}

G.K. wrote and conceived the manuscript with support from A.C. and R.F.
G.K. is author of half of the functions implemented in the \proglang{R}
package \pkg{abn}. G.K. performed the numerical simulations and
contributed to the analysis. G.K wrote the package's documentation.
F.I.L. is the original creator and author of half of the functions in
the \proglang{R} package \pkg{abn}. F.I.L. provided critical feedback
and helped to shape the research project. A.C. identified the case study
dataset, performed the analysis and contributed to the interpretation of
the findings. M.P. contributed to the \proglang{R} package \pkg{abn}
documentation. R.F. is the PhD supervisor of G.K. and M.P. R.F. probed
the \proglang{R} package \pkg{abn} and provided useful suggestions that
lead to numerous improvements on the package. R.F. provided input on the
statistical framework, model implementation and findings reporting. All
authors revised the manuscript.

\hypertarget{computational-details}{%
\section*{Computational details}\label{computational-details}}
\addcontentsline{toc}{section}{Computational details}

The results in this paper were obtained using \proglang{R} 3.6.1
(2019-07-05) -- ``Action of the Toes'' with the \pkg{abn} 2.2 package.
All computations were carried out using RStudio 1.2.5001. \proglang{R}
and all packages used are available from the Comprehensive \proglang{R}
Archive Network (CRAN) at \url{https://CRAN.R-project.org/}. RStudio is
available at \url{https://www.rstudio.com/}. The \proglang{JAGS} version
is 4.3.0. The \proglang{INLA} version is 3.6.

\appendix

\hypertarget{sec:dagsim}{%
\section{DAGs simulation: more technical details}\label{sec:dagsim}}

In order to simulate DAGs, the function \code{simulateDag()} generates
triangular matrices with user tuneble arc density. The node ordering is
sampled from the node definition.

\begin{figure}[!ht]
  \centering
\includegraphics[width=\textwidth]{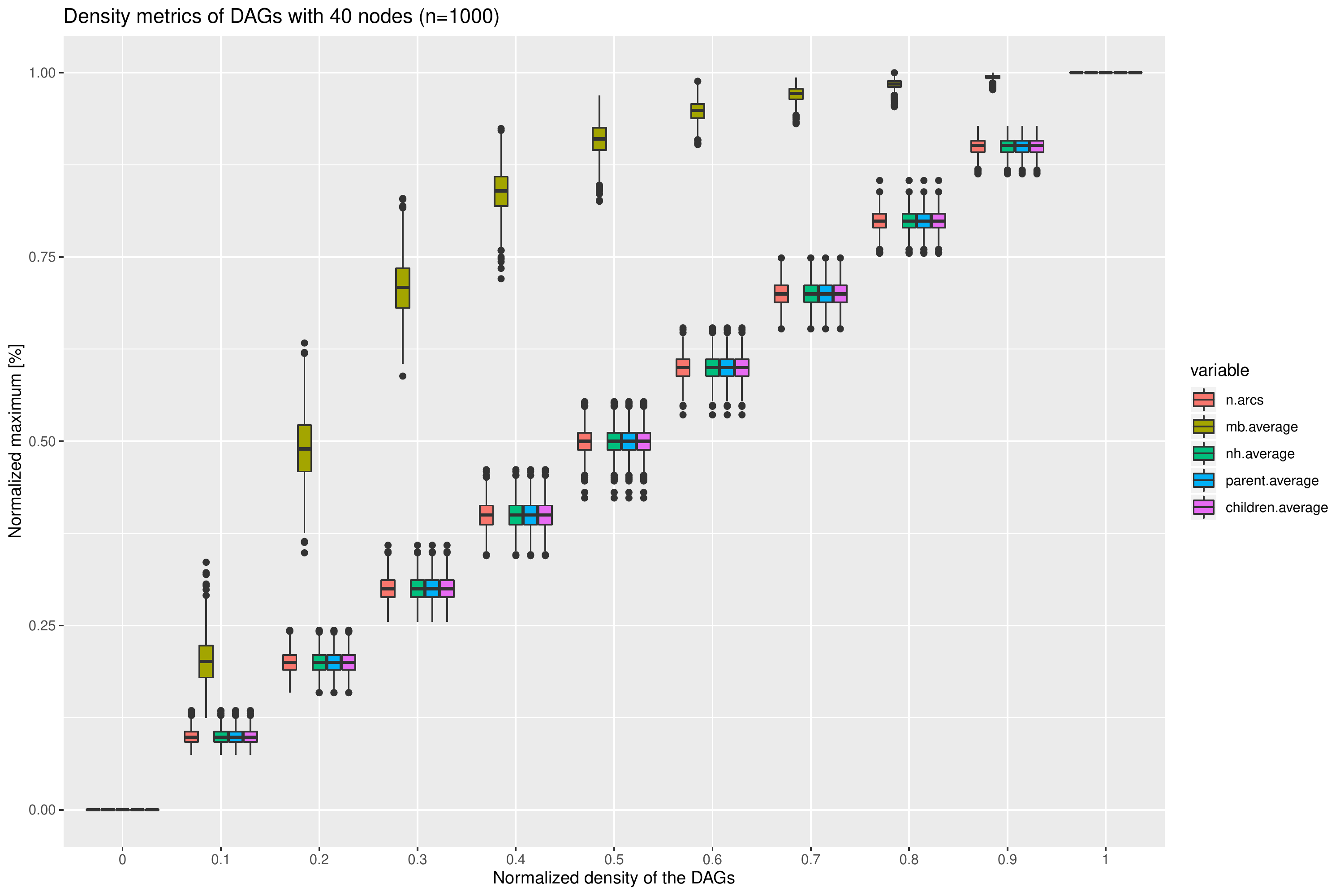}
  \caption{Normalized DAG structural metrics from simulated Bayesian networks of 40 nodes with different arc densities}
  \label{fig:dag_40}
\end{figure}

In Figure \ref{fig:dag_40}, some normalized (i.e., divided by the
maximum possible) DAGs structural metrics are displayed: n.arcs is the
number of arcs, mb.average is the average size of the Markov blanket
(the set of parents, children and co-parents), nh.average is the average
number of neighbours, parent.average is the average number of parents
and children.average is the average number of children in the BN. For
each level of network complexity, 1000 BN with 40 nodes were simulated.
The normalized distribution of the BN metrics computed with
\code{infDag()} is reported. As one can see in Figure \ref{fig:dag_40},
the only normalized metric that exhibits non-linear behaviour with arc
density is the Markov blanket.

\hypertarget{subsec:coefestim}{%
\section{Regression coefficients estimation: quality assurance
check}\label{subsec:coefestim}}

\begin{figure}[!ht]
  \centering
\includegraphics[width=\textwidth]{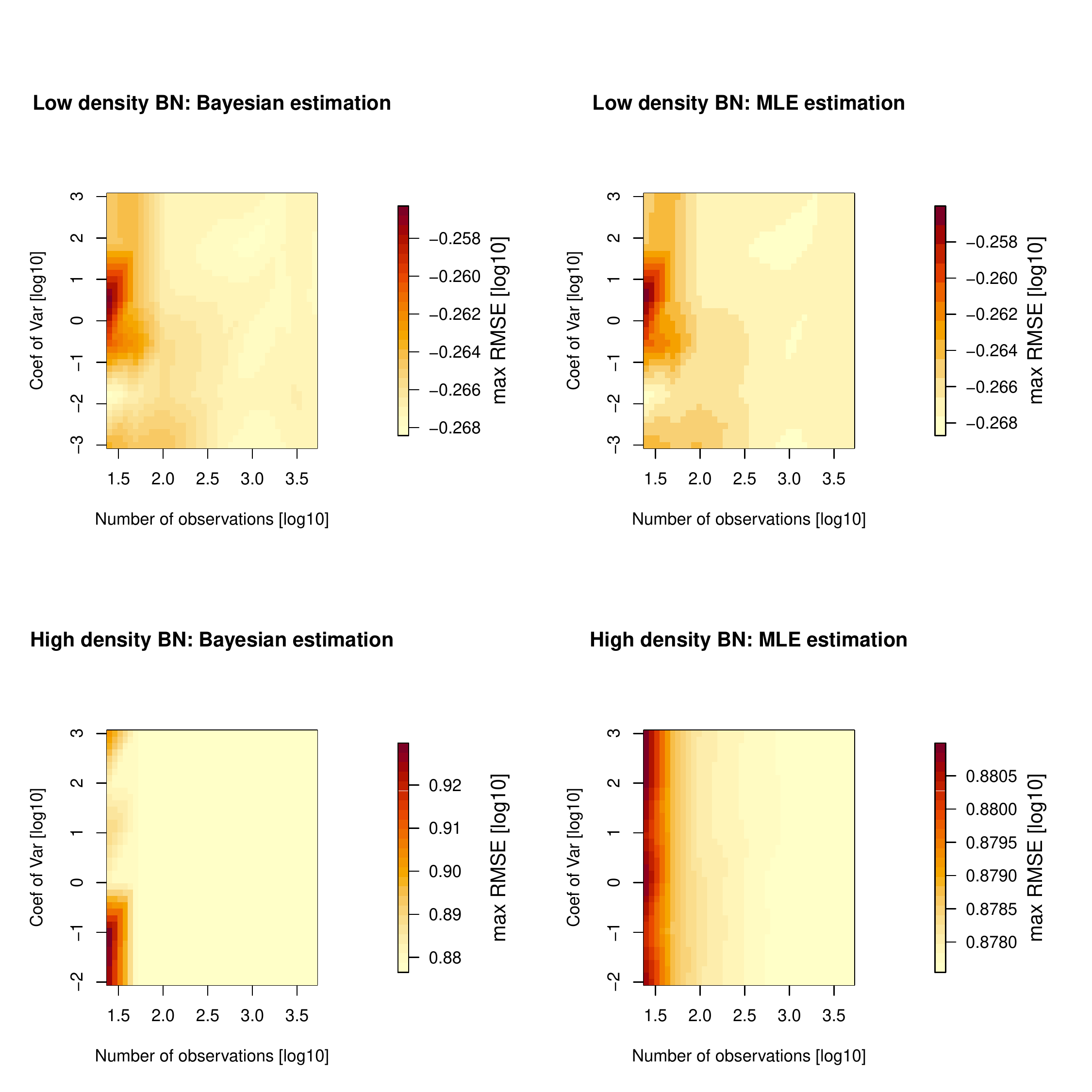}
  \caption{Comparison between Bayesian and MLE implementations to estimate regression coefficients in an ABN framework. The panels show the maximum Root Mean Squared Error (max RMSE) in function of the network density, the distribution variability and the sample size.}
  \label{fig:arc_estimation}
\end{figure}

The \proglang{R} package \pkg{abn} contains two functions to estimate
the regression coefficent based on two statistical paradigms: the MLE
formulation with an IRLS estimation, and a Bayesian formulation with
diffuse priors estimated with \proglang{INLA}. Figure
\ref{fig:arc_estimation} shows a quality assurance check of the
implementations.

The Bayesian and MLE implementations are compared in Figure
\ref{fig:arc_estimation} to estimate the accuracy of parameters. Two
network densities, 20\% and 80\% of the possible arcs expressed, were
simulated 50 times. Then the regression coefficients were computed for
different sample sizes and the coefficient of variation of the given
node as a proxy for the distribution's variability. As one can see in
Figure \ref{fig:arc_estimation}, the error is measured as the maximum
Root Mean Squared Error (max RMSE) on a log-log scale. Both
implementations produce very similar results even if the estimation
frameworks are very different.

\hypertarget{sec:paramboot}{%
\section{Parametric bootstrapping: the R code}\label{sec:paramboot}}

This section contains the \proglang{R} code needed to perform parametric
bootstrapping described by the pseudo-code in Algorithm
\ref{alg:pseudoboot}.

\begin{CodeChunk}

\begin{CodeInput}
R> vars <- colnames(abndata)
R> 
R> # load data for jags
R> source("PostParams.R")
R> 
R> # select nr. bootstrap samples to run
R> set.seed(123)
R> 
R> # get 5000 random numbers to set different initial values
R> n <- sample(1:100000, 5000)
R> 
R> # specify max number of parents based on previous search
R> max.par <- 4
R> 
R> # Simulate data and run ABN on such dataset
R> for (i in 1:length(n)) {
+      print(paste("/n Running simulation", i))
+      # pick initials
+      init <- list(.RNG.name = "base::Mersenne-Twister", .RNG.seed = n[i])
+      # run model
+      jj <- jags.model(file = "plot/model8vPois.bug", data = list(AR.p = AR.p, 
+          pneumS.p = pneumS.p, female.p = female.p, livdam.p = livdam.p, 
+          eggs.p = eggs.p, wormCount.p = wormCount.p, age.p = age.p, 
+          prec.age.p = prec.age.p, adg.p = adg.p, prec.adg.p = prec.adg.p), 
+          inits = init, n.chains = 1, n.adapt = 5000)
+      # run more iterations
+      update(jj, 100000)
+      # sample data (same size as original: 341) with a sampling
+      # lag (20) to reduce autocorrelation
+      samp <- coda.samples(jj, c("AR", "pneumS", "female", "livdam", 
+          "eggs", "wormCount", "age", "prec.age", "adg", "prec.adg"), 
+          n.iter = n.obs * 20, thin = 20)
+      # build dataframe in the same shape as the original one
+      dt.boot <- as.data.frame(as.matrix(samp))
+      dt.boot <- dt.boot[, vars]
+      # now coerce to factors if need be and set levels
+      abndata <- as.data.frame(abndata)
+      for (j in 1:length(vars)) {
+          if (is.factor(abndata[, j])) {
+              dt.boot[, j] <- as.factor(dt.boot[, j])
+              levels(dt.boot[, j]) <- levels(abndata[, j])
+          }
+      }
+      # Build a cache of all local computations
+      mycache <- buildscorecache(data.df = dt.boot, data.dists = dist, 
+          dag.banned = banned, dag.retained = retain, max.parents = max.par)
+      # Run the EXACT SEARCH
+      mp.dag <- mostprobable(score.cache = mycache)
+      fabn <- fitabn(object = mp.dag)
+      # Save the results obtained
+      save(mycache, mp.dag, fabn, dt.boot, file = sprintf("boot.%04d.RData", 
+          i))
+ }
\end{CodeInput}
\end{CodeChunk}

\hypertarget{numerical-stability-check-parameter-estimation}{%
\section{Numerical stability check: Parameter
estimation}\label{numerical-stability-check-parameter-estimation}}

Once the trimmed DAG is obtained to keep robust structural features, we
can extract the marginal posterior densities. Indeed, a BN has a
qualitative part (the structure) and a quantitative part (parameters
estimates). They are both equally important for interpreting and
reporting the findings. As the estimation is based on the Laplace
approximation and the \proglang{R} package \pkg{abn} does not rely on
conjugate priors, a numerical check is performed. Figure
\ref{fig:mp_plot} shows the marginal densities. This figure is much more
ambitious than the structural learning phase. The critical assumption is
that the data contains sufficient information to accurately estimate the
density of individual parameter of the model. This is a stronger
requirement than simply being able to estimate an overall goodness of
fit metric. It could happen that \proglang{INLA} or the internal
\proglang{C} code cannot estimate the densities with enough numerical
stability. Many user-tunable parameters can be supplie to
\code{fitabn()} by providing a list to the \code{control} argument.

\begin{CodeChunk}
\begin{figure}[h]

{\centering \includegraphics{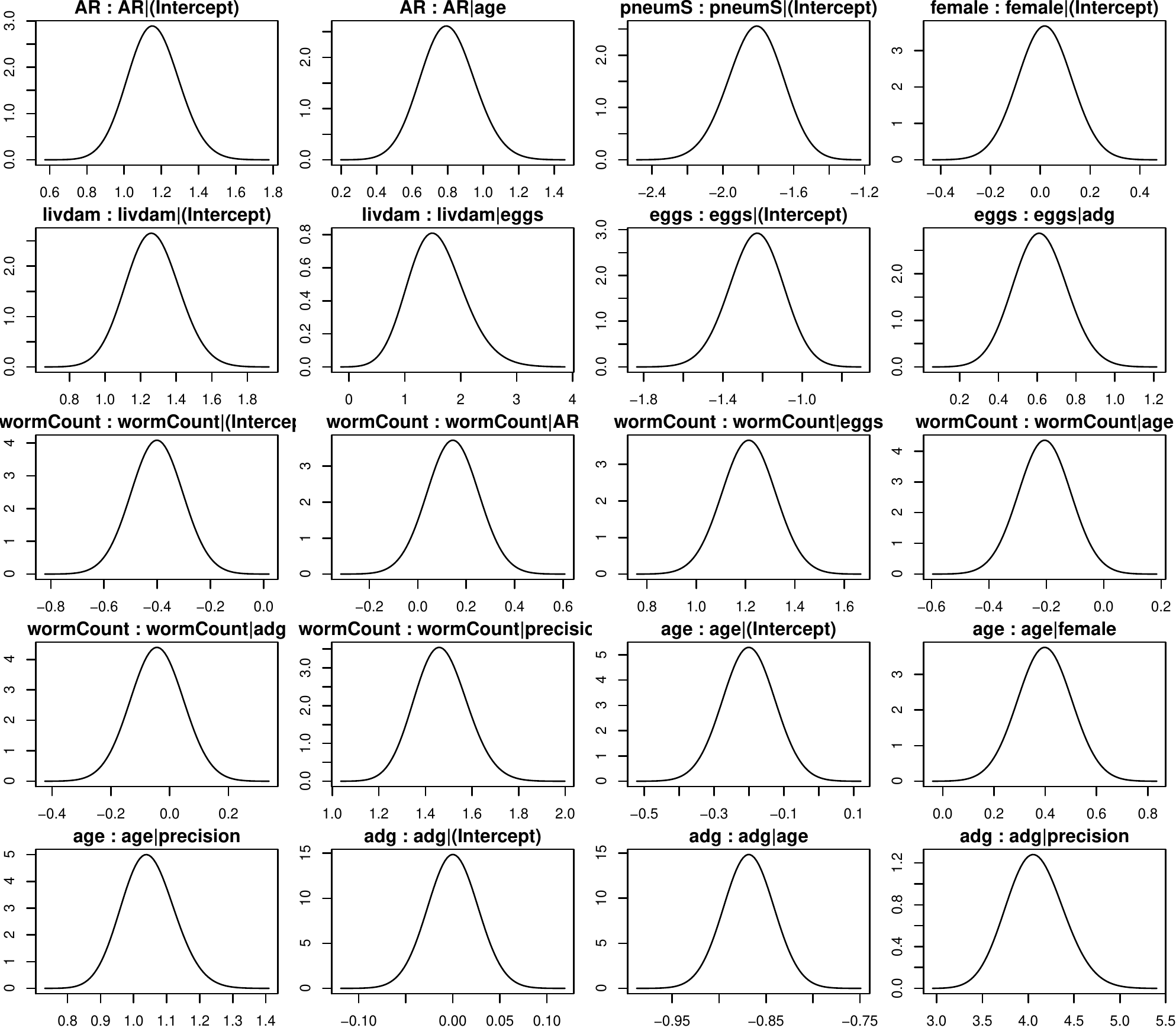} 

}

\caption{\label{fig:mp_plot}Marginal densities of model parameters}\label{fig:mpd_plot}
\end{figure}
\end{CodeChunk}

Both the shape of the posterior parameter distributions and their area
under the density look satisfactory (see Figures \ref{fig:mp_plot} and
\ref{fig:auc}).

\begin{CodeChunk}
\begin{figure}[h]

{\centering \includegraphics{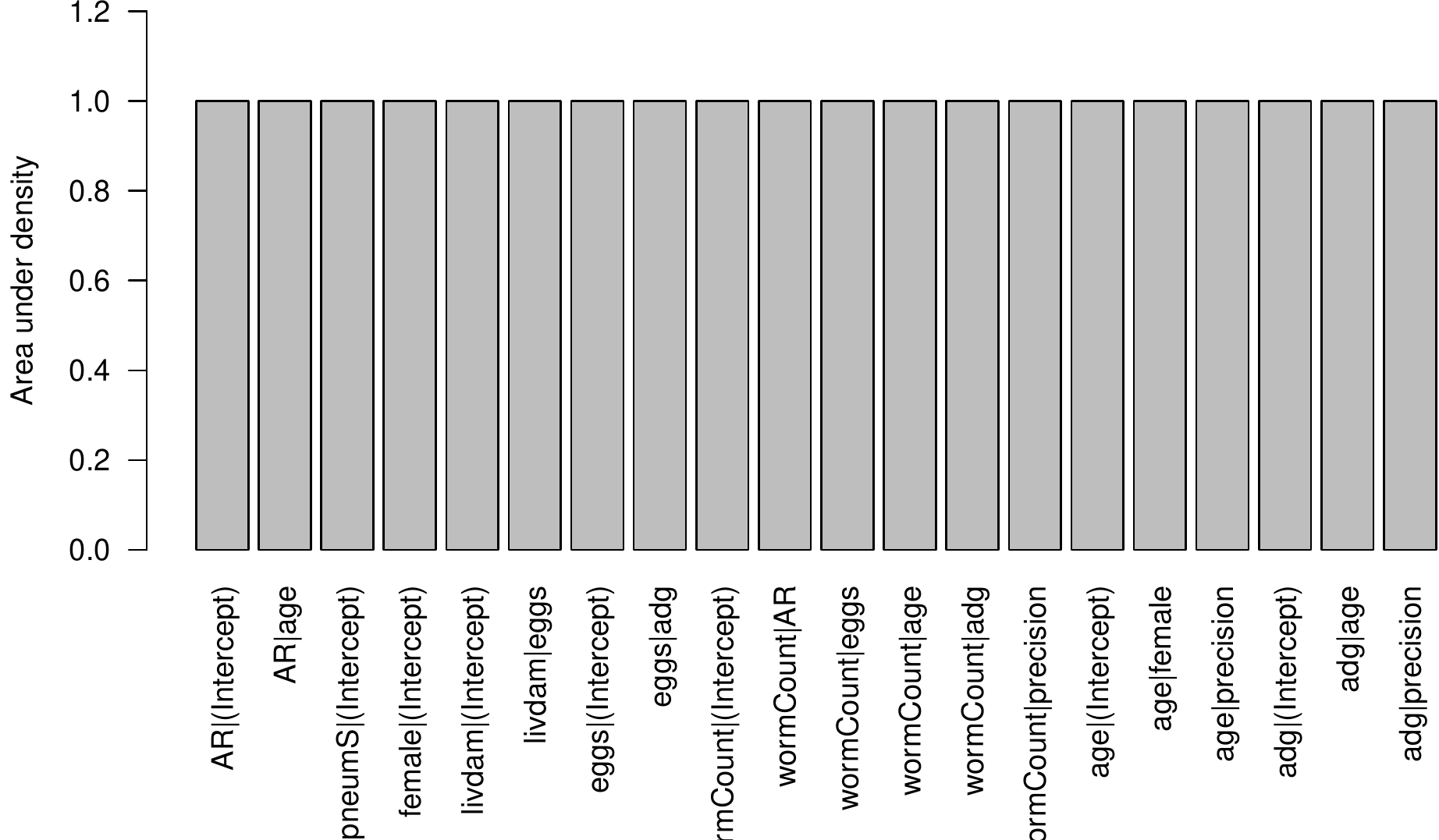} 

}

\caption{\label{fig:auc} Area under the posterior density for model parameters.}\label{fig:mpd_auc}
\end{figure}
\end{CodeChunk}

\bibliography{refs.bib}

\end{document}